\documentclass[twoside]{article}

\usepackage[accepted]{aistats2022}
\usepackage[round]{natbib}
\usepackage[utf8]{inputenc}
\usepackage{amssymb}
\usepackage{amsmath}
\usepackage{amsthm}
\usepackage{graphicx}
\usepackage{nicefrac}
\usepackage{pifont}
\usepackage{tikz}
\usepackage{thmtools}
\usepackage{thm-restate}
\usepackage{xcolor}
\usepackage{cellspace}

\newcommand{\fsum}{f_{\textnormal{add}}}

\newcommand{\EXb}[1]{\mathbb{E}_{\mathbf{X}}\left[ #1 \right]}

\newcommand{\EYgivenXb}[1]{\mathbb{E}_{Y|\mathbf{X}}\left[ #1 \right]}
\newcommand{\EYsb}[1]{\mathbb{E}_{Y}\left[ #1 \right]}
\newcommand{\EXYsb}[1]{\mathbb{E}_{\mathbf{X}Y}\left[ #1 \right]}
\newcommand{\Eygivenx}{\mathbb{E}_{Y\mid {\bf X=x}}}

\newcommand{\EDb}[1]{\mathbb{E}_{\scriptstyle D}\left[#1 \right]}
\newcommand{\Egenb}[2]{\mathbb{E}_{#1} \left[ #2 \right]}{}

\DeclareMathOperator*{\argmin}{\arg\!\min}

\newcommand{\qbar}{\overline{q}}

\newcommand{\fbar}{\overline{f}}

\newcommand{\vectorx}{\mathbf{x}}

\newcommand{\vectorX}{\mathbf{X}}

\newcommand{\Yzeroone}{Y_{0/1}}
\newcommand{\sign}{\textnormal{sgn}}

\newcommand\averagei{ \frac{1}{M}\sum_{i=1}^M }

\newcommand{\gen}[0]{\ensuremath{\phi}}
\newcommand{\Bregman}[2]{\ensuremath{B_{\gen}(#1, #2)}}
\newcommand{\BregmanGen}[3]{\ensuremath{B_{#1}\left(#2, #3\right)}}

\newcommand{\optimallink}{\boldsymbol{\psi}}
\newcommand{\minrisk}{\underline{L}}

\newcommand{\margin}{\ell}

\newcommand{\prob}{\mathbb{P}}
\newcommand{\cmark}{\ding{51}}%
\newcommand{\xmark}{\ding{55}}%

\newcommand{\fhat}{\widehat{f}}

\newcommand{\gpos}{g^+}
\newcommand{\gneg}{g^-}

\newcommand{\limpos}{\lim_{g \rightarrow \gpos}}
\newcommand{\limneg}{\lim_{g \rightarrow \gneg}}

\newtheorem{theorem}{Theorem}
\newtheorem{lemma}{Lemma}
\newtheorem{corollary}{Corollary}
\newtheorem{proposition}{Proposition}
\newtheorem{definition}{Definition}

\begin{document}



\twocolumn[

\aistatstitle{Bias-Variance Decompositions for Margin  Losses}

\aistatsauthor{ Danny Wood \And Tingting Mu \And  Gavin Brown }

\aistatsaddress{ University of Manchester\And  University of Manchester \And University of Manchester } ]

\begin{abstract}
    We introduce a novel bias-variance decomposition for a  range of strictly convex margin losses, including the logistic loss (minimized by the classic {\em LogitBoost} algorithm), as well as the squared margin loss and canonical boosting loss.
    Furthermore, we show that, for {\em all} strictly convex margin losses,
    the expected risk decomposes into the risk of a ``central'' model and a term quantifying variation in the functional margin with respect to variations in the training data.
    These decompositions provide a diagnostic tool for practitioners to understand model overfitting/underfitting,
    and have implications for additive ensemble models---for example, when our bias-variance decomposition holds, there is a corresponding ``ambiguity'' decomposition, which can be used to quantify model diversity.
    
\end{abstract}

\section{INTRODUCTION}

Bias-variance decompositions are broadly recognized as an important tool for understanding the generalization properties of statistical models~\citep{Geman1992, Belkin2018}. The most well-known of these decompositions is for the squared loss~\citep{Geman1992}, but such decompositions have been shown to hold for other losses,
including KL-divergences between 
probability densities~\citep{Heskes1998, Hansen2000}, proper scoring rules~\citep{Buja2005}
and Bregman divergences~\citep{Pfau2013}. These decompositions
allow us to see how much of our model's generalization error is due to poor model selection
versus being due to randomness in the training data/learning procedure~\citep{Geman1992, Belkin2018, Adlam2020}.

In this work, we consider bias-variance decompositions for margin losses---i.e., surrogate loss functions for binary classification, where the task of classification is re-framed as estimation of a real value,  with larger magnitudes corresponding to more confident predictions of the positive/negative class.
Margin losses are used in classical algorithms and models such as AdaBoost~\citep{Freund1999}, LogitBoost~\citep{Friedman2000}, and Support Vector Machines (SVMs), yet the issue of bias-variance decompositions for these losses has been largely overlooked. To the best of our knowledge, the only applicable previous work is~\cite{Buja2005}, which converts the losses into proper scoring rules. Their framework requires re-interpretation the real-valued model as a probability estimator---an undesirable intermediate step.   

We present an alternative approach, not dependent on interpreting the models as probability estimators. 
The resulting decomposition is applicable to a broad class of losses which we characterize for the first time and refer to as {\em gradient-symmetric losses}.
We also show that this family of losses is of independent interest, having strong connections with canonical form losses \citep{Masnadi2010}, linear odd losses \citep{Patrini2016}, and Bregman divergences. Furthermore, in the context of ensemble methods, we show that gradient-symmetric losses are exactly the sub-class of margin losses where we obtain an ambiguity decomposition \citep{Krogh1995} for linearly combined ensembles. A full list of our contributions is as follows.

\begin{itemize}
    \item We derive a novel decomposition of the expected risk applicable to \emph{any} strictly convex margin loss, separating the contributions of the expected model and variation in the functional margin.
   \item We introduce \emph{gradient-symmetric} losses, a class of margin losses including the logistic, squared, Laplacian~\citep{Masnadi2011} and canonical boosting losses~\citep{Masnadi2010}. For these losses, we present a novel bias-variance decomposition, which does not require interpreting models as probability estimators.% without the need to re-map models to probability estimates.
    \item For gradient-symmetric losses, we derive an ensemble ambiguity decomposition \citep{Krogh1995} for linearly combined ensembles.  For non-gradient-symmetric losses, we show a similar decomposition, but requiring a {\em non-linear} ensemble combination rule.

    \item We show a close relationship between gradient-symmetric losses and Bregman divergences: gradient-symmetric losses are expressible in terms of a Bregman divergence from a representation of the label to the model.
    
    \item We examine how gradient-symmetric losses relate to canonical form losses~\citep{Masnadi2010} and linear odd losses~\citep{Patrini2016}. Using the latter connection, we show that the bias term in our decomposition can be decomposed further into an expected margin term and a target-independent term---suggesting possible applications to semi-supervised settings. 
\end{itemize}

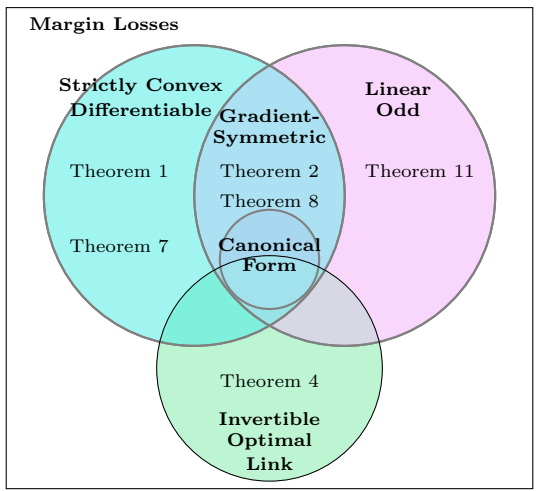
\begin{figure}\label{fig:venn}
\centering
% Set A and B
 \definecolor{babyblueeyes2}{HTML}{bcd4e6}

\definecolor{gradsymcolor}{HTML}{43ebe4}
\definecolor{diffcolor}{HTML}{c5dedd}
\definecolor{linoddcolor}{HTML}{f3aff9}
\definecolor{optcolor}{HTML}{7ceba6}
\definecolor{cancolor}{HTML}{b6c1e2}

% \definecolor{gradsymcolor}{HTML}{3a7c89}
% \definecolor{diffcolor}{HTML}{c5dedd}
% \definecolor{linoddcolor}{HTML}{ffa679}
% \definecolor{optcolor}{HTML}{f16d7a}
% \definecolor{cancolor}{HTML}{ffa679}

 \def\firstcircle{(0,0) circle (2cm)}
 \def\secondcircle{(0:2cm) circle (2cm)}
 \def\thirdcircle{(1cm,-.85cm) circle (0.66cm)}
 \def\fourthcircle{(1cm, -2.3cm) circle (1.5cm)}

  \colorlet{circle edge}{black!50}
  \colorlet{circle area}{blue!20}

  \tikzset{filled/.style={fill=circle area, draw=circle edge, thick}, outline/.style={draw=circle edge, thick}}

  \setlength{\parskip}{5mm}
 \begin{tikzpicture}
%  
%      \begin{scope}
%       \clip \firstcircle;
%       \fill[gradsymcolor] \secondcircle;
%     \end{scope}
%     
%     \fill[babyblueeyes2] \thirdcircle;
%  
%  
%     \begin{scope}[even odd rule]% first circle without the second
%         \clip \secondcircle (-2.5,-2.5) rectangle (4.2,2.5);
%     \fill[diffcolor] \firstcircle;
%     \end{scope}
%     
%     \begin{scope}[even odd rule]% first circle without the second
%         \clip \firstcircle (-2.5,-2.5) rectangle (4.2,2.5);
%     \fill[linoddcolor] \secondcircle;
%     \end{scope}
    \begin{scope}[blend group = soft light]
        \fill[gradsymcolor, opacity=.5] \firstcircle; 
        \fill[linoddcolor, opacity=.5] \secondcircle; 
        \fill[optcolor, opacity=.5] \fourthcircle; 
        \fill[cancolor] \thirdcircle; 
    \end{scope}

  \begin{scope}
    %\clip \secondcircle;
    \draw[outline] \firstcircle
                                 \secondcircle node [label={[xshift=0.71cm, yshift=0.8cm]{\scriptsize\bf\shortstack{Linear\\ Odd}}}] {};
  \end{scope}
     \draw[outline] \firstcircle node [label={[xshift=-0.71cm, yshift=0.8cm]{\scriptsize\bf\shortstack{Strictly Convex\\ Differentiable}}}] {}
               \secondcircle;

   \draw[outline] \thirdcircle {} node [label={[yshift=-12] {\scriptsize\bf\shortstack{Canonical\\Form}}}] {};

    \draw \fourthcircle {} node [label={[xshift=-.0cm, yshift=-1.6cm] {\scriptsize\bf\shortstack{Invertible\\Optimal\\Link}}}] {};

   \draw (-2.5,-3.9) rectangle (4.5,2.5) node [text=black,above] {};
   \draw (1, 0) node [label={[yshift=.4cm]\scriptsize\bf\shortstack{Gradient-\\Symmetric}}]{};
  
   \draw (1, 0) node [label=\scriptsize Theorem~\ref{the:f_bv}]{};
   \draw (-1, 0) node [label=\scriptsize Theorem~\ref{the:margin_variance_decomposition}]{};
   
   \draw (3, 0) node [label=\scriptsize{Theorem~\ref{the:lol_decomp}}]{};
   
   \draw (1, -.4) node [label=\scriptsize{Theorem~\ref{the:grad_sym_ambig}}]{};
%   \draw (3, -1) node [label=\scriptsize \eqref{eq:lol_bias}+\eqref{eq:lol_variance}]{};
   
   \draw (-1, -1) node [label=\scriptsize Theorem~\ref{the:general_ambiguity} ]{};
   
   \draw (1, -2.8) node [label=\scriptsize Theorem~\ref{the:buja_bv} ]{};
   
   \draw (-1.2, 1.9) node [label=\bf\scriptsize{Margin Losses} ]{};

   \end{tikzpicture}
   \caption{Visualization of How Our Contributions Relate to Known Classes of Margin Loss.}
   \end{figure}

The applicability of our results to various known classes of margin loss is shown in Figure~\ref{fig:venn}.
All the decompositions we present have a similar flavour: each separates the error into  the risk of a ``central" model and a stochastic component. The ``central" model is the model obtained by integrating out all sources of randomness in the training data and learning algorithm,
and the risk of this model is the sum of a systematic component and the error due to noise in the label distribution.

\section{PRELIMINARIES}

%\subsection{Margin Losses}
%
{\noindent\bf Margin Losses }
We consider the following learning scenario: We have the problem of learning a function $h: \mathcal{X} \rightarrow \{-1, +1\}$, which minimizes the zero-one loss $\frac{1}{N}\sum_{j=1}^N [[h(\vectorx_j) \neq y_j]]$ for a training set $\mathcal{D} =\{(\vectorx_j, y_j)\}_{j=1}^N$ with feature vector $\vectorx_j \in \mathcal{X}$ and target $y_j \in \{-1, +1\}$. 
We consider margin losses, i.e., surrogate losses of the form $\ell(yf(\vectorx; \mathcal{D}))$, where $\ell: \mathbb{R} \rightarrow \mathbb{R}_+$ is an upper bound on the zero-one loss and $f(\cdot\,;\,\cdot)$ 
is the model, which is a function of the input $\vectorx \in \mathcal{X}$ and parameterized by the training set $\mathcal{D}$.
We use the model to predict a label $y \in \{-1, +1\}$ for $\vectorx \in \mathcal{X}$ by mapping the model output to a class prediction with the rule $h(\vectorx) = \textnormal{sign}(f(\vectorx; \mathcal{D}))$.
Note that these losses have the property that the loss associated with an incorrect prediction is not affected by the class labelling scheme, i.e., $\ell((-y)(-f(\vectorx; \mathcal{D})))) = \ell(yf(\vectorx; \mathcal{D}))$.

The value $yf(\vectorx; \mathcal{D})$ is known as the \emph{functional margin}; positive values correspond to correct classifications, while negative values correspond to incorrect classifications. The magnitude of this margin can be interpreted as
%corresponding to
the model's confidence in its prediction, and has been associated with good generalization properties via bounds on the error~\citep{Schapire1998}.

We will assume throughout this paper that the infimum of $\ell$ is zero. Most results can easily be modified for other finite lower bounds, but it is usually only sensible to consider losses where the minimizer(s) of $\ell$ are positive.

%\subsection{Bregman Divergences}

\begin{figure}
    \centering
    \includegraphics[width=.45\textwidth]{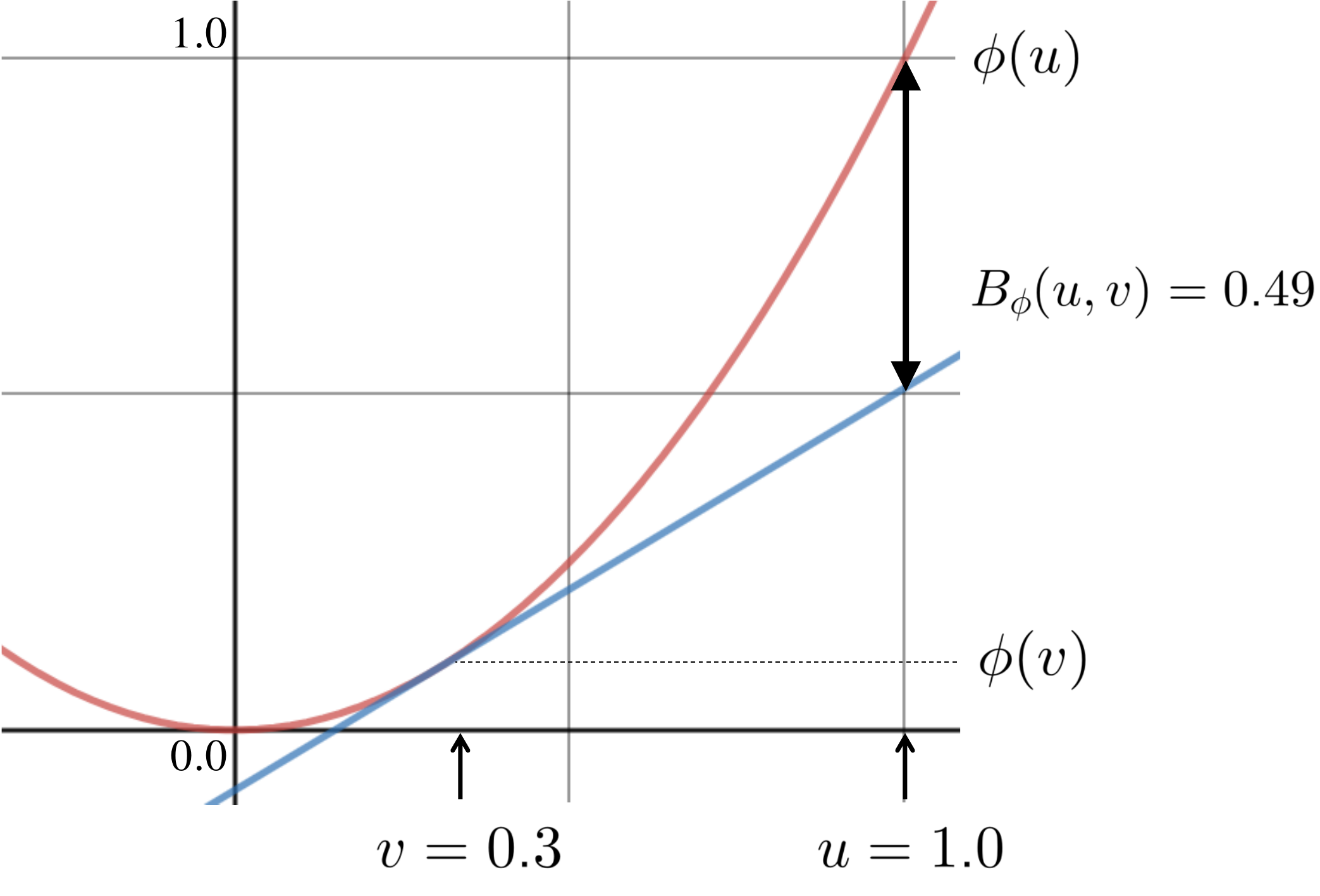}
    \caption{The Bregman divergence, illustrated for the Bregman generator $\phi(u)=u^2$.} % For the example shown, this results in a divergence $\Bregman{u}{v}=(u-v)^2=(1.0-0.3)^2 = 0.49$}
    \label{fig:bregman}
\end{figure}

{\noindent\bf Bregman Divergences }
Bregman divergences~\citep{Bregman1967} are measures of separation between points, defined in terms of a strictly convex function $\phi$, and are a frequently used tool in the study of margin losses~\citep{Zhang2004,Masnadi2008, Nock2008bregman, Reid2010}.
Loosely speaking, the Bregman divergence $\Bregman{u}{v}$ is the difference between the strictly convex function $\phi$ evaluated at $u$, and the linear approximation of $\phi$ around $v$, also evaluated at $u$ (See Figure~\ref{fig:bregman}).

More formally, for some convex set $S$, the Bregman divergence  $B_\phi: S \times \textnormal{int}(S) \rightarrow \mathbb{R}_+$ is defined as 
\begin{align*}
    \Bregman{u}{v} = \phi(u) - \phi(v) - \phi'(v)(u-v),
\end{align*}
where $\textnormal{int}(S)$ denotes the interior of the set $S$ and $\phi$ is a strictly convex differentiable function. Bregman divergences are not generally symmetric, i.e., it does not necessarily hold that $\BregmanGen{\phi}{u}{v}=\BregmanGen{\phi}{v}{u}$.

%\subsection{Bias-Variance Decompositions}
{\noindent\bf Bias-Variance Decompositions }
In the typical supervised binary classification learning setup, a model is trained on a set of training examples $\mathcal{D} = \{(\vectorx_j, y_j)\}_{j=1}^N$, where each example is drawn independently from a joint distribution over $\mathcal{X} \times \{-1, +1\}$. It is therefore natural, when considering the impact of varying the training set on model performance, to consider the training set as a random variable $D=\{(\vectorX_j, Y_j)\}_{j=1}^N$, where each $(\vectorX_j, Y_j)$ is independently distributed according to the aforementioned joint distribution.
We consider the performance of the model $f(\cdot; D)$ over variations in $D$ 
and over the true distribution of the data, which we write as the joint distribution of $(\vectorX, Y)$. 
When we consider the random variable $f(\vectorX; D)$, we will typically leave the arguments implicit, simply writing $f$. We also note that though $\EDb{\cdot}$ is the expectation over the distribution of training sets, the decomposition can just as easily apply for any source of stochasticity in the model or training data (e.g., randomization of initial weights). 

For the squared loss, the setup is the same, except that $Y \in \mathbb{R}$. Here, the
classic
%classic bias-variance
decomposition of~\citep{Geman1992} tells us that the expected risk can be decomposed as
\begin{align*}
    \underbrace{\EXYsb{\EDb{(f-Y)^2}}}_{\textnormal{expected risk}} &= \underbrace{\EXYsb{(Y - Y^\ast)^2 }}_\textnormal{noise}+\\ &\hspace{-1.1cm} \underbrace{\EXb{(Y^\ast - f^\ast)^2}}_{\textnormal{bias}} +  \underbrace{\EXb{\EDb{(f^\ast -f)^2}}}_\textnormal{variance},  
\end{align*}
where $Y^\ast= \EYgivenXb{Y}$ and $f^\ast = \EDb{f}$.
The bias gives a measure of the distance of the ``central" model $f^\ast$ from the expected target, while the variance quantifies how far the typical model will be from this central model.

These quantities can be estimated in practice by training multiple models on different training sets, approximating the expectations with averages over the multiple datasets. This can either be done via training on disjoint subsets of the training set~\citep{Yang2020} or by training on bootstraps of the training data~\citep{Neal2018}.

Other known bias-variance decompositions have the same structure, separating the expected risk into noise, bias and variance, with the variance term measuring the spread of models in the distribution around some central model (for example, see ~\cite{Pfau2013,Heskes1998}).

Bias-variance decompositions bring important insights: a large systematic component (i.e., high bias) 
suggests that the model is insufficiently complex (and therefore underfitting) whereas a large stochastic component (i.e., high variance) suggests over-sensitivity to the particular random sample (e.g., overfitting to training data).

\section{BIAS-VARIANCE DECOMPOSITION FOR MARGIN LOSSES}

\subsection{Decomposition for Margin Losses}\label{sec:bias_variance}
We are interested in the expected risk (or expected generalization error) of the model; for margin losses, this is defined as $\EXYsb{\EDb{\ell(Yf(\vectorX; D)}}$. 

We begin with a general decomposition which has some of the properties we are looking for: it contains a term which measures the loss of a ``central" model and a term which depends on how spread out the distribution of models is around this centre. However, the latter term is \emph{not}, in general, independent of the target $Y$.

\begin{restatable}[Margin Variance Decomposition]{theorem}{generalBV}\label{the:margin_variance_decomposition}
For a strictly convex differentiable margin loss $\ell$,
     \begin{align}
        \underbrace{\EXYsb{\EDb{\ell(Y f)}}}_{\textnormal{expected risk}} =& \underbrace{\EXYsb{\ell(Y f^\ast)}}_{\textnormal{risk of central model}} +\nonumber \\
         & \hspace{.5cm} \underbrace{\EXYsb{\EDb{\BregmanGen{\ell}{Y f}{Y f^\ast}}}}_{\textnormal{margin variance}},\label{eq:gen_bv}
    \end{align}   
    where we define the central model $f^\ast = \EDb{f}$. 
\end{restatable}

Proof of this result, and all other results presented in this paper, can be found in the appendices.

The results shows that we can decompose the expected risk of the model into the loss of the expected model and a non-negative term 
quantifying the spread of the models around that expectation. We refer to this decomposition as the \emph{margin variance decomposition} because for $y \in \{-1, +1\}$, the term $\EDb{\BregmanGen{\ell}{yf}{yf^\ast}}$ measures the spread of the functional margin around the expectation $\EDb{yf}=yf^\ast$.

\begin{figure}[ht]
    \centering
    \includegraphics[width=.46\textwidth]{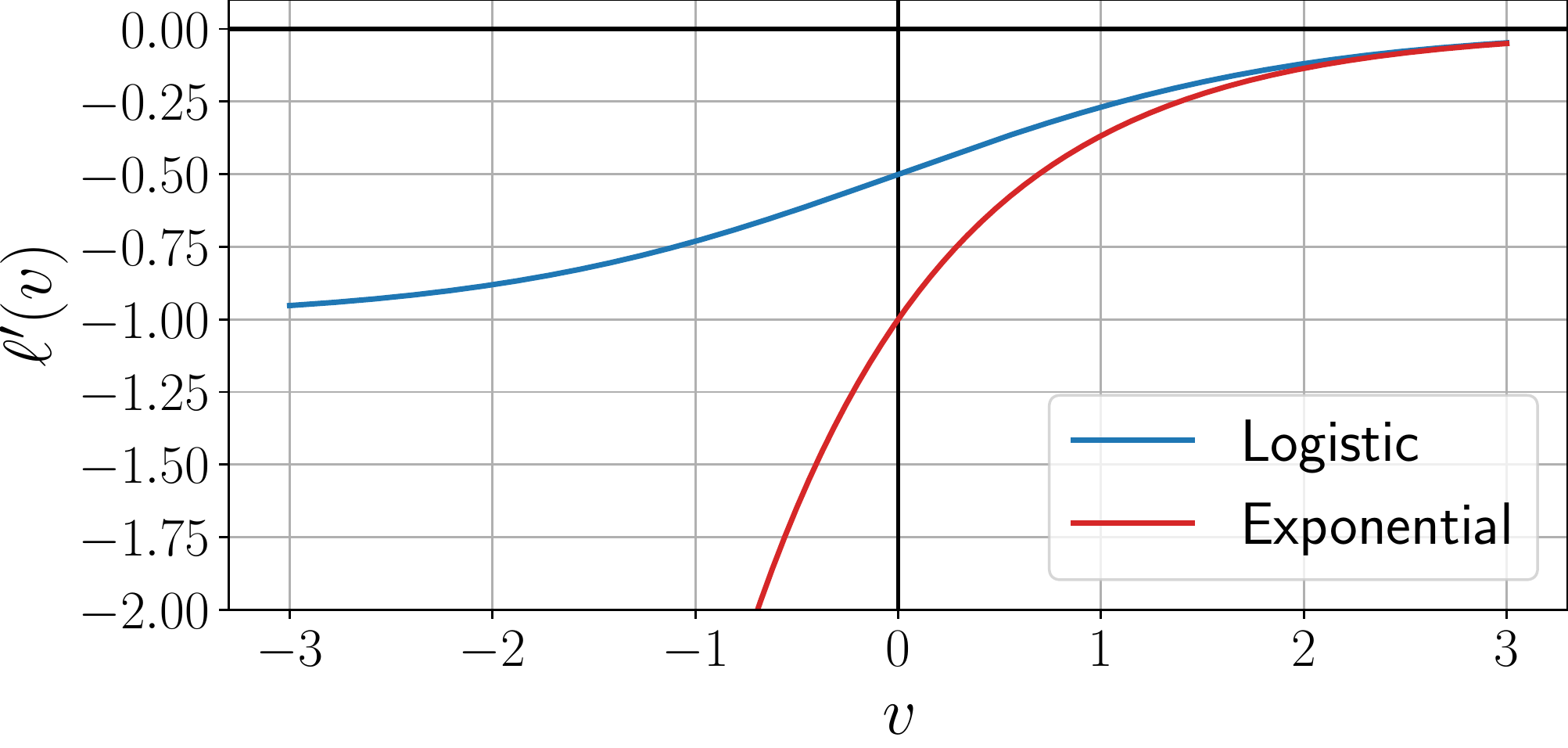}
    \caption{Comparison of Gradients of Logistic and Exponential Losses. The Logistic Gradient is Rotationally Symmetric About a Point on the y-axis, Hence we Refer to the Logistic Loss as a Gradient-Symmetric Loss.}\label{fig:grad_sym_losses}
\end{figure}

Even with this target-dependent variance term, this decomposition gives insight into how much of the error is due the systematic part of the model (the central model $f^\ast$) versus the random part (the distribution of $f$).
However, there is a class of loss where the decomposition becomes a true bias-variance decomposition with a target-independent variance term. We consider the class of margin losses with the following property.
\begin{definition}[Gradient-Symmetry]
    Let $\ell: \mathbb{R} \rightarrow \mathbb{R}_+$ be a differentiable strictly-convex function. $\ell$ is gradient-symmetric if for all $v \in \mathbb{R}$
    \begin{align}
        \ell'(v) + \ell'(-v)=c \label{eq:grad_condition}
    \end{align}
    for some constant $c$.
\end{definition}
Examples of gradient-symmetric losses (and notable non-gradient-symmetric ones) are given in Table~\ref{tab:losses}, and
illustrated
in Figure~\ref{fig:grad_sym_losses}. 
For practical purposes, we are only interested in losses where $c<0$, since losses with $c>0$ have the undesirable property of penalizing models more harshly for correct predictions than incorrect ones and losses with $c=0$ ignore the sign of the label. With this restriction, gradient-symmetric losses are closely related to canonical form losses---each such gradient-symmetric loss is, up to a constant scaling factor, equivalent to a canonical form loss.
However, the gradient-symmetry definition remains useful, as it completely captures the class of losses for which the margin-variance in~\eqref{eq:gen_bv} becomes target-independent. This is due to %gradient-symmetric losses having 
the following property.
\begin{restatable}{proposition}{gradientSymmetricProperty}
    \label{prop:grad_sym_condition}
    For all $u, v \in \mathbb{R}$ and strictly convex differentiable $\ell: \mathbb{R} \rightarrow \mathbb{R}_+$,
    \begin{align*}
        \BregmanGen{\ell}{u}{v} = \BregmanGen{\ell}{-u}{-v}
    \end{align*}
    if and only if $\ell$ is gradient-symmetric.
\end{restatable}

This result means that for gradient-symmetric losses, the sign of $Y$ in $\BregmanGen{\ell}{Yf}{Yf^\ast}$ has no effect. Applying this fact to the margin variance decomposition, we get the following bias-variance decomposition. 

\begin{restatable}[Bias-Variance Decomposition for Gradient-Symmetric Losses]{theorem}{gradientSymmetricBV}\label{the:f_bv}
    For a strictly convex differentiable margin loss $\ell$, there exists a bias-variance decomposition of the form
    \begin{align}
        \underbrace{\EXYsb{\EDb{\ell(Yf)}}}_{\textnormal{expected risk}} &= \underbrace{\EXYsb{\ell(Yf^\ast)}}_\textnormal{noise + bias} + \underbrace{\EXb{\EDb{\BregmanGen{\ell}{f}{f^\ast}}}}_\textnormal{variance}~\label{eq:grad_sym_bv}
    \end{align}
    if and only if $\ell$ is gradient-symmetric.
\end{restatable}

It is also possible to separate out the noise and bias (see Appendix~\ref{sec:noise}), but from a practical standpoint combining the two terms into a single expression is often more useful, since we do not usually know the distribution of the class labels and therefore do not have enough information to disentangle the contributions to the expected risk from the bias and the noise. 

Finally, we note that it is possible to relax the condition that the loss is differentiable. If instead, we just require an injective mapping between points and their subgradients, we can extend the definitions of Bregman divergences and gradient-symmetry accordingly to retain the decompositions we have introduced.

\begin{table*}[ht]
    \centering\def\arraystretch{1.8}
    \begin{tabular}{c|c|c|c|c}
    {\bf Name} 
        & \bf\shortstack{ Loss\\ $\ell(v)$}
        & \bf\shortstack{Gradient \\ $\ell'(v)$}  
        & \bf \shortstack{Gradient-\\Symmetric?} 
        & \bf \shortstack{Constant\\$c$} \\
        %& \bf \shortstack{Classification/\\Regression Loss}\\
        \hline\hline
    \small Squared  
        & $(1-v)^2$
        & $2(v-1)$   
        &  \cmark
        & -4\\
        %& Regression\\
    \small Logistic 
        &  $\log(1+ \exp(-v))$
        & $\displaystyle -\frac{1}{1 + \exp(v)}$
        & \cmark
        & -1 \\
        %& Classification\\
    \small \shortstack{Canonical Boosting\\ Loss} %\citep{Masnadi2010}}
        &  $\frac{1}{2}\sqrt{v^2 + 4} - \frac{1}{2}v $
        & $ \displaystyle \frac{v}{2\sqrt{v^2 + 4}} - \frac{1}{2}$
        & \cmark
        & -1 \\
        %& Classification \\
    \small Laplacian Loss %\citep{Masnadi2015}
        &  $\frac{1}{2} \left( \exp(-|v|) + |v| - v \right)$
        & $-\frac{1}{2} + \frac{1}{2} \left( \sign(v)(1 - \exp(-|v|)) \right)$ 
        & \cmark
        & -1\\
        %& Classification\\
    \small Exponential
        &  $\exp(-v)$
        & $-\exp(-v)$ 
        & \xmark
        & N/A\\
        %& Classification\\
         
         \small Smooth Hinge 
         & $\frac{1}{t}\ln(1 + \exp(-t (v-1)))$
         & $\displaystyle - \frac{1}{1 + \exp(t (v-1))}$
         & \xmark  
         & N/A\\ \hline
         %& ? \\
         %& Classification\\

    \end{tabular}
    \caption{Examples of Margin Loss Functions and Their Derivatives}
    \label{tab:losses}
\end{table*}

\subsection{Comparison to Buja's Decomposition}\label{sec:bv_proper}

\cite{Buja2005} proposed a bias-variance decomposition for proper scoring rules, and also showed its application to margin losses via re-interpretation of the model as a probability estimator. This intermediate step requires a {\em link} function to perform the mapping, as well as the introduction of several other concepts.
%derived via examining the gradient of the minimum risk. 
We have shown that gradient-symmetric losses do not require this intermediate step, instead having a bias-variance decomposition in terms of the raw model output, $f$.  
In order to compare the merits of the two decompositions, we now examine the properties of \cite{Buja2005}'s decomposition.

In order to present this decomposition, we must first introduce some new concepts and notation. 
Let $p=\prob(Y=+1| \vectorX)$, i.e., $p$ is the probability the target is $+1$ given  $\vectorX$, then $L(p, f)$, the expected  pointwise risk of a model $f$ when $\vectorX = \vectorx$,  can be written 
\begin{align*}
    L(p, f) = \Eygivenx{\ell(Yf)} = p \ell(f) + (1-p) \ell(-f).
\end{align*}
We define the minimum risk, $\minrisk(p)$ as the infimum of $L(p,f)$ for a fixed $p$, that is,
\begin{align*}
    \minrisk(p) &= \inf_{v \in \mathbb{R}}L(p, v) = \inf_{v \in \mathbb{R}} p \ell(v) + (1-p) \ell(-v).
\end{align*}
We also define the optimal link function, $\psi(p)$ as mapping from $p\in (0,1)$ to the $v \in \mathbb{R}$ at which the minimum is achieved, i.e.,
\begin{align*}
    \psi(p) &= \argmin_{v \in \mathbb{R}} L(p, v) = \argmin_{v \in \mathbb{R}} p \ell(v) + (1-p) \ell(-v).
\end{align*}
Assuming this function is invertible, we may think of the model $f$ as implicitly defining a probability and write this probability as $q= \psi^{-1}(f)$.  

Up until now, we have thought of $f$ as being our model, with our class prediction being determined by the sign of $f$. If instead we think of our model as $q$, we can reframe the task as estimating the conditional probability of the positive class given the feature vector $\vectorx$ (we could make the dependence of $q$ on $\vectorx$ and the training set $\mathcal{D}$ explicit by writing $q(\vectorx; \mathcal{D})$). Using the optimal link function $\psi$, we may now consider the loss $\ell(y \psi(q))$. This loss has the property of being a strictly proper scoring rule, i.e., $L(p, \psi(q))$ is minimized if and only if $q=p$. Here, Bregman divergences can be used to measure the separation between the desired probability estimate $p$ and the one obtained from the model, $q$. This Bregman divergence measures the \emph{excess risk}, i.e., the risk of a model minus the amount of irreducible error due to label noise. 
\begin{theorem}(\citep{Zhang2004})\label{the:zhang}
    When $\psi$ is invertible and differentiable, the pointwise excess risk can be written as a Bregman divergence of the form
    \begin{align}
        L(p, \psi(q)) - \minrisk(p) = \BregmanGen{-\minrisk}{p}{q}. \label{eq:excess_bregman}
    \end{align}
\end{theorem}

With this result, we are able to construct a bias-variance decomposition for the excess risk. This can be transformed into a bias-variance decomposition on the risk itself by noting that the minimum risk for losses with infimum zero losses is the same as the noise.

\begin{restatable}[Buja's Bias-Variance Decomposition]{theorem}{bujaBV}\label{the:buja_bv}
For loss $\ell$ with invertible differentiable link function $\psi$,
    \begin{align*}
    \underbrace{\EXb{\EDb{L(p, f)} - \minrisk(p)} }_{\textnormal{expected excess risk}}
    %&=  \EXYb{\EDb{\BregmanGen{-\minrisk}{\Yzeroone}{\,q^\ast}}} + \EXb{\EDb{\BregmanGen{-\minrisk}{q^\ast}{q}}}\\ 
    &=  \underbrace{\EXb{\BregmanGen{-\minrisk}{p}{\,q^\ast}}}_\textnormal{bias}\\
    & \hspace{1cm}+ \underbrace{\EXb{\EDb{\BregmanGen{-\minrisk}{q^\ast}{q}}}}_\textnormal{variance}
    %&= \underbrace{\EXYb{\ell(Y q^\ast)}}_\textnormal{bias} + \underbrace{\EXb{\EDb{\BregmanGen{-\minrisk}{q^\ast}{q}}}}_\textnormal{variance},
    \end{align*}
    where $q^\ast = [-\minrisk']^{-1} \left(\EDb{-\minrisk'(q)}\right)$. %Equivalently,
%     \begin{align*}
%         \EXYsb{\EDb{\ell(Yf}} =& \underbrace{\EXb{\minrisk(p)}}_\textnormal{noise} + 
%         \underbrace{\EXb{\EDb{\BregmanGen{-\minrisk}{p}{\,q^\ast}}}}_\textnormal{bias}\\
%         &\hspace{1.5cm} + \underbrace{\EXb{\EDb{\BregmanGen{-\minrisk}{q^\ast}{q}}}}_\textnormal{variance}.
%     \end{align*}
\end{restatable}
A version of this decomposition was first noted in~\citep{Buja2005}, though we formulate it here using notation and terminology from \citep{Zhang2004} and \citep{Reid2010}. Note that this decomposition can be applied for loss functions where the link function is invertible, including non-gradient-symmetric losses such as the exponential loss which minimized by AdaBoost~\citep{Friedman2000}.

For the case of the logistic loss, the minimum risk is $-p \ln p - (1-p) \ln (1-p)$. This is the binary entropy, and in this case $\BregmanGen{-\minrisk}{\cdot}{\cdot}$ becomes the KL-divergence. With this connection, we see that Buja's decomposition is in fact a special case of the decomposition for probability distributions discussed in~\citep{Heskes1998}. 

We will shortly examine how Theorem~\ref{the:buja_bv} connects to Theorem~\ref{the:f_bv}, but comparing it with the more general decomposition in Theorem~\ref{the:margin_variance_decomposition} we find: 
\begin{itemize}
\item Theorem~\ref{the:margin_variance_decomposition} allows a more intuitive choice of ``central" model $f^\ast = \EDb{f}$ which, as shown later, enables its use in a wider range of ensemble  models, but its variance term is only target-independent under the gradient-symmetric condition.
\item Although the variance term in Theorem~\ref{the:buja_bv} is target-independent, it requires a particular form for the ``central" probability prediction $q^\ast$, which does not necessarily correspond to the prediction of $f^\ast$. In particular, $q^\ast$ is defined to satisfy
\begin{align*}
    q^\ast = \argmin_{\widehat{q} \in (0,1) }\EDb{\BregmanGen{-\minrisk}{\widehat{q}}{q}}.
\end{align*}
\item These two models coincide (i.e., $f^\ast =\psi(q^\ast)$) if and only if $\ell$ is gradient-symmetric. In this case, the two decompositions are equivalent whenever the conditions for both are met. This is stated more formally in Theorem~\ref{the:coinciding_centroids} in Appendix \ref{app:coinciding_centroids}.
\item Though Theorem~\ref{the:buja_bv} does not require gradient-symmetry, for some losses (such as the squared loss), the requirement for invertible $\psi$ can necessitate restricting permitted model outputs. For example, in the case of the squared loss, $\psi(p) = 2p -1$, so the range of $\psi$ is $[-1, 1]$. 

\end{itemize}

The strength of Buja's decomposition is that it turns the loss into a proper composite loss via the inverse link function~\citep{Reid2010}. These objects have since been studied in detail, both for binary classification and the multi-class case~\citep{Vernet2011}. However, there are some cases, e.g., boosting, where though the models are excellent classifier, they do \emph{not} give good probability estimates~\citep{Mease2008}. This makes it somewhat unnatural to apply Buja's decomposition in these cases. 

\subsection{Connections Between the Two Decompositions}\label{sec:proper}

This paper has discussed two bias-variance decompositions for gradient-symmetric losses; it is natural to ask how these two are related.
We can show a deep connection between the two decompositions in Theorem~\ref{the:f_bv} and Theorem~\ref{the:buja_bv} via the notion of \emph{dual} Bregman divergences.
Every Bregman divergence has an equivalent dual formulation defined by the convex conjugate of the generator~\citep{Banerjee2005}. Let $[\phi]^\ast$ be the convex conjugate of the differentiable strictly convex function $\phi: S \rightarrow \mathbb{R}$, then for all $u, v \in \textnormal{int}(S)$,
\begin{align*}
   \Bregman{u}{v}  = \BregmanGen{[\phi]^\ast}{\phi'(v)}{\phi'(u)}.
\end{align*}
Note that, for the dual formulation, we require \emph{both} arguments be on the interior of their domain. 

Given that we have two Bregman divergences of interest, $B_\ell$ and $B_{-\minrisk}$, one may wonder whether they are related by this duality. 
Indeed they are, but the relationship is not quite as straightforward as saying that $\ell$ and $-\minrisk$ are convex conjugates of each other; there is the presence of the additional scaling factor $c$. The actual relationship is described by the following result.

\begin{restatable}{theorem}{negativeLoss}\label{prop:negative_loss}
    Let $\ell$ be a gradient-symmetric margin loss with negative minimum risk $-\minrisk$, then
    \begin{align*}
        [-\minrisk]^\ast(v) = \ell\left(\frac{v}{c}\right).
    \end{align*}
%     and equivalently
%     \begin{align*}
%         [-\minrisk]^\ast(cf) = \ell\left(f\right),
%     \end{align*}
%     In particular, for canonical form losses
%     \begin{align*}
%         [-\minrisk]^\ast(f) = \ell\left(-f\right).
%     \end{align*}
\end{restatable}
Examples of this relationship can be observed in the last two columns of Table~\ref{tab:convex_conjugates}. 
With this, we can show that for gradient-symmetric losses, the two decompositions are linked by the following result.
\begin{restatable}{theorem}{dualTheorem}\label{the:dual_connection}
    For strictly convex invertible gradient-symmetric loss $\ell$ with infimum zero and with invertible $-\minrisk'$, for all $u, v \in \mathbb{R}$,
    \begin{align*}
         \BregmanGen{\ell}{u}{v} = \BregmanGen{-\minrisk}{[-\minrisk']^{-1}\left(cv\right)}{[-\minrisk']^{-1}\left(cu \right)}.
    \end{align*}
\end{restatable}
Hence the two decompositions are intimately linked by Bregman duality, even though the Bregman divergences are not strictly duals of each other. It is interesting to note that the connection could equally be derived via the \emph{matching loss} approach of~\cite{Helmbold1995}. When conditions for both decompositions are met, i.e., for gradient-symmetric losses with a differentiable invertible link, \emph{the variance terms in the two decompositions are exactly equivalent} (see Appendix~\ref{app:equiv}). 

\setlength\cellspacetoplimit{5pt}
\begin{table*}[ht]
    \centering\def\arraystretch{1.8}
    \begin{tabular}{c|Sc|c|c}
    {\bf Name} 
        %& \bf\shortstack{Optimal Link \\ $f=\optimallink(q)$}  
        & \bf\shortstack{Negative\\Minimum Risk\\ $-\minrisk(p)$}
        & \bf\shortstack{Conjugate Negative\\ Minimum Risk \\ $[-\minrisk]^\ast(v)$}  
        & \bf \shortstack{Loss function\\ $\ell(v)$} \\
        %& \shortstack{\bf Minimum risk\\ $\minrisk(p)$}  
        \hline\hline
    \small Squared  
        & $-4p(1-p)$
        %& $2q-1$   
        & $\left(1+\frac{v}{4}\right)^2$   
        &  $(1-v)^2$\\
        %& $4p(1-p)$ \\
    \small Logistic 
        &  $p \ln p + (1-p) \ln (1-p)$
        %& $\frac{1}{2}\ln\frac{q}{1-q}$   
        & $\ln(1 + \exp(v))$
        & $\ln(1 + \exp(-v))$\\
        %& $2\sqrt{p(1-p)}$ \\
    \small Canonical Boosting Loss 
        &  $-2 \sqrt{p(1-p)}$
        %& $\ln\frac{q}{1-q}$   
        & $\frac{1}{2} \sqrt{p^2 + 4} + \frac{1}{2}p$ 
        %& \small $-p\ln p-(1-p)\ln(1-p)$
        & $\frac{1}{2} \sqrt{v^2 + 4} - \frac{1}{2}v$\\
    \small Laplacian Loss %\citep{Masnadi2011}
        & \small\shortstack{$-\frac{1}{2}(1 - |2p-1|)\cdot\hphantom{\ln 2p-1}$\\ $(1- \ln (1 - |2p-1|))$}\normalsize 
        %& $\ln\frac{q}{1-q}$   
        & \small$\frac{1}{2} \left( \exp(-|v|) + |v| + v \right)$\normalsize 
        %& \small $-p\ln p-(1-p)\ln(1-p)$
        & \small$\frac{1}{2} \left( \exp(-|v|) + |v| - v \right)$\normalsize\\
    %Savage & $\frac{1}{(1+e^{yf})^2}$ & $\ln\frac{q}{1-q}$ &  $p(1-p)$ 
    \end{tabular}
    \caption{Canonical Form Losses Along With Their Negative Minimum Risks and Corresponding Conjugates %\dw{Laplacian min risk looks like $2 \sign(p- 1/2)(|p-1/2| + (2p-2) \ln (2 \sign(p-1/2)|p-1/2| )) $. Maybe leave this row out of this table? } 
    }
    \label{tab:convex_conjugates}
\end{table*}

\section{AMBIGUITY DECOMPOSITION FOR MARGIN LOSSES}

It is a well-known phenomenon in machine learning that ensembles of estimators tend to outperform individual models. For ensembles of regression models, this notion is formalized by the \emph{ambiguity decomposition}~\citep{Krogh1995}, which shows that, under the squared loss, the error of the average of distinct model outputs is guaranteed to be less than the average error of the individuals. More formally, for $M$ model outputs, $f_1, \ldots, f_M$, combined by taking the arithmetic mean $\fbar = \averagei f_i$,
for the label $y$,
\begin{align*}
   \underbrace{(\fbar - y)^2}_\textnormal{ensemble error} = \underbrace{\averagei (y - f_i)^2}_\textnormal{average error} - \underbrace{\averagei (\fbar - f_i)^2}_\textnormal{ambiguity}. 
\end{align*}
The ambiguity is a measure of ensemble diversity, independent of the target $y$.
This decomposition can be thought of as a rearrangement of a special case of the bias-variance decomposition, taking the average over a finite set of models rather than expectation over $D$.

\subsection{Deriving an Ambiguity Decomposition}\label{sec:ambiguity}
%\subsection{\textcolor{red}{Decomposition By  Theorem~\ref{the:margin_variance_decomposition}  }}\label{sec:ambiguity}

We derive an ambiguity decomposition for margin losses following the same approach as Section~\ref{sec:bias_variance}, giving a more general decomposition in terms of the margin, then showing for gradient-symmetric losses, the ambiguity term becomes target-independent.
\begin{restatable}[Margin Ambiguity Decomposition]{theorem}{generalAmbiguity}\label{the:general_ambiguity}
    Let $\ell$ be a strictly convex differentiable margin loss and $\fbar = \averagei f_i$ be an ensemble of models, then
    \begin{align*}
        \ell(y\fbar)
        &= \underbrace{\averagei \ell(y f_i)}_{\textnormal{average error}} - \underbrace{\averagei \BregmanGen{\ell}{yf_i}{y\fbar}}_\textnormal{margin ambiguity}.%~\label{eq:gen_ambiguity}
    \end{align*}

\end{restatable}
\begin{restatable}[Ambiguity Decomposition for \mbox{Gradient}-Symmetric Margin Losses]{theorem}{gradientSymmetricAmbiguity}\label{the:grad_sym_ambig}
    With the same setup as in Theorem~\ref{the:general_ambiguity}, we have the ambiguity decomposition
    %Let $\ell$ be a strictly convex margin loss, then we have the ambiguity decomposition
    \begin{align*}
        \ell(y\fbar) %\lim_{g \rightarrow g_y} \averagei \BregmanGen{\ell}{f_i}{g_y} - \averagei \BregmanGen{\ell}{f_i}{\fbar}\\
        &= \underbrace{\averagei \ell(yf_i)}_\textnormal{average error} - \underbrace{\averagei \BregmanGen{\ell}{f_i}{\fbar}}_\textnormal{ambiguity}.
    \end{align*}
    if and only if $\ell$ is gradient-symmetric.
\end{restatable}

This gives the existence of a simple ambiguity decomposition for the range of losses discussed in the previous section.
Margin losses are frequently used with additive ensembles, like those constructed by LogitBoost, where ensemble members are combined by summation rather than averaging \citep{Friedman2000}. For these cases we can modify Theorem~\ref{the:grad_sym_ambig} by re-formulating the ensemble output as an average $\fsum=\sum_{i=1}^M f_i=\averagei M f_i$. 

\begin{corollary}\label{cor:ambig}
    Let $f_1, \ldots, f_M$ be $M$ model outputs and ensemble output $\fsum = \sum_i f_i$, then 
    \begin{align*}
        \ell(y\fsum) = \averagei \ell(y M f_i)  - \averagei \BregmanGen{\ell}{M f_i}{~\fsum}.
    \end{align*}
\end{corollary}
The same trick generalizes to arbitrary weighted combination rules using $\sum_i \alpha_i f_i = \frac{1}{M} \sum_i \alpha_i M f_i$.

\subsection{Ambiguity via Buja's Decomposition}
%\subsection{\textcolor{red}{Discussion on Use of Theorem~\ref{the:buja_bv}}}
One might expect that we can find an ambiguity decomposition for non-gradient-symmetric losses via proper scoring rules, in the same way that Theorem~\ref{the:buja_bv} gives a bias-variance decomposition for those losses. However, for linearly combined models, this is not the case. When $\qbar$ is the centroid combiner $\qbar = [-\minrisk']^{-1}\left( \averagei -\minrisk'(q_i) \right)$, we have
\begin{align}
    \BregmanGen{-\minrisk}{p}{\qbar} = \averagei \BregmanGen{-\minrisk}{p}{q_i}  - \averagei \BregmanGen{-\minrisk}{\qbar}{q_i} \label{eq:buja_ambiguity};
\end{align}
this has the form required for an ambiguity decomposition, but crucially, \emph{only holds for the centroid combiner $\qbar$}. The decomposition would therefore require that $\fbar = \psi(\qbar)$. Equivalently, from the definition of $\qbar$, we need
\begin{align}
    \fbar = \psi \left([-\minrisk']^{-1}\left( \averagei -\minrisk'(\psi^{-1}(f_i)) \right)\right).~\label{eq:gen_mean}
\end{align}

In order for this to be the case, we need a specific condition to hold, as described by the following theorem.
\begin{restatable}{theorem}{combinerIsLinear}
    Let $\ell$ be a differentiable loss function with a continuous invertible  link function, then
    \begin{align}
        \psi^{-1}(\fbar) = [-\minrisk']^{-1}\averagei[-\minrisk]'(\psi^{-1}(f_i))
    \end{align}
    if and only if 
     $\psi(p) = -a \minrisk'(p)$
    for some constant $a \neq 0$.
    \end{restatable}
This condition turns out to be equivalent to gradient-symmetry, giving the following corollary.
\begin{restatable}{corollary}{ambiguityCondition}
    Let $f_1, \ldots, f_M$ be an ensemble of models with ensemble output $\fbar$ a linear combination of the ensemble members. A target-independent ambiguity decomposition of form in~\eqref{eq:buja_ambiguity} exists for linearly combined models if and only if $\ell$ is gradient-symmetric. 
\end{restatable}

When constructing new ensemble methods,~\eqref{eq:gen_mean} provides a natural combination rule for a given loss that guarantees that an ambiguity decomposition exists for a given ensemble. We leave exploring the merits of this rule in ensemble learning to future work.

\section{CONNECTIONS TO EXISTING WORK}\label{sec:grad_sym}

Gradient-symmetric losses are closely related to several other notable classes of functions. In this section, we highlight these connections and their implications:

\begin{itemize}
\item They are exactly the class of losses such that $\ell(yf)$ is  a Bregman divergence where the first argument of the divergence is the raw model output $f$ (or potentially the limit of such a divergence).
\item They are a superset of canonical form losses~\citep{Masnadi2010}.
\item They are a subset of linear-odd losses~\citep{Patrini2016}, for which we derive a novel decomposition of the expected risk.
\end{itemize}

\subsection{Connection to Bregman Divergences}

Proposition~\ref{prop:grad_sym_condition} explains the bias-variance decomposition for gradient-symmetric losses as a property of the Bregman divergence $B_\ell$. Here, we show this is due to a deeper connection: for gradient-symmetric losses, the loss $\ell(yf)$ itself is expressible directly in terms of $B_\ell$.

\begin{restatable}[Gradient-Symmetric Losses as Bregman Divergences]{theorem}{gsAreBregman}\label{the:grad_sym_breg}
 Let $\ell: \mathbb{R} \rightarrow \mathbb{R}$ be a strictly-convex  differentiable margin loss function with infimum zero, then $\ell(yf)$ is expressible as (the limit of) a Bregman divergence with $f$ as the first argument if and only if $\ell$ is gradient-symmetric.
 Furthermore, if $\ell(yf)$ is expressible as such a Bregman divergence, it is of the form
 \begin{align}
     \ell(yf) = \lim_{g \rightarrow g_y} \BregmanGen{\ell}{f}{g}. \label{eq:conv_bregman}
 \end{align}
 where $g_y$ is a constant determined by $y$. 
\end{restatable}
For losses where $\lim_{v \rightarrow \infty} \ell(v)=0$, $g_y \in \{-\infty, \infty\}$; otherwise, the values of $g_y$ are finite and the $B_\ell(f, g_y)$ can be evaluated directly. 
Note that in the former case, 
$\lim_{v\rightarrow\infty} \ell'(v)=0$, and therefore $\lim_{v\rightarrow\infty} \ell'(-v)=c$, so the gradient tends to a constant in both directions.

A consequence of Theorem~\ref{the:grad_sym_breg} is that where $\ell$ appears in our decompositions, it can be replaced with an expression of the form in the right-hand side of \eqref{eq:conv_bregman}. Doing so, we find our decomposition to be closely related to the decompositions for Bregman divergences shown in~\citep{Pfau2013}, despite arising from a very different setting.

\subsection{Connection to Canonical Form Losses}

Canonical form losses \citep{Masnadi2010} (also referred to as \emph{permissible convex losses} \citep{Nock2008bregman}) are a family of losses with  a tight coupling between the minimum risk and the model's interpretation as a probability estimator. In particular, a loss function $\ell$ is said to be in canonical form if the derivative of the negative minimum risk function coincides with the link function, i.e, 
    $-\minrisk'(p)= \psi(p)$
for all $p \in (0, 1)$. This condition was shown in \citep{Masnadi2011} to be equivalent to 
\begin{align}
    \ell'(v) + \ell'(-v) = -1. ~\label{eq:canonical_condition}
\end{align}
Comparing~\eqref{eq:grad_condition} and~\eqref{eq:canonical_condition}, gradient-symmetric loss functions can be seen to be a more general class of losses, since the value of $c$ is not restricted to $-1$. 
However, they are not a great deal more general: every gradient-symmetric loss with $c< 0$ can be scaled by a constant factor to become a canonical loss. 
More formally:
\begin{restatable}{proposition}{gradientSymmetricOptimal}\label{prop:grad_sym_optimal_link}
    A differentiable strictly convex margin loss $\ell$ is gradient-symmetric with constant $c<0$ if and only if it satisfies
    \begin{align}
        \minrisk'(p)= c \psi(p) \label{eq:well_behaved_link}
    \end{align}
Furthermore, the loss $\frac{1}{|c|}\ell$ is a canonical form loss.
\end{restatable}

Canonical form losses are also the class of losses for which $\BregmanGen{\ell}{u}{v}=\BregmanGen{[-\minrisk]^\ast}{u}{v}$, so in this case Bregman divergences \emph{are} duals of each other, though \emph{not} because $\ell$ are $-\minrisk$ convex conjugates. Instead, $-\minrisk^\ast(v)= \ell(-v)$ and the equivalence is due to gradient symmetry implying that $\ell(-v) = \ell(v) -v$ and Bregman generators differing by only affine terms forming equivalence classes.

\subsection{Connection to Linear Odd Losses}

The Linear Odd Losses (LOLs) are an important class of loss, introduced in \citep{Patrini2016} and characterized by the fact that they factor into a linear term and a target-independent term.
This structure makes these losses of particular interest in weakly supervised and semi-supervised learning settings.

While gradient-symmetric losses are a superset of canonical form losses, they are a \emph{subset} of the LOLs, as described in~\citep{Patrini2016}.
As the name suggests, a LOL is a loss where the odd part is linear, as in the following definition. 
\begin{definition}[Linear Odd Loss]
    Any loss function $\ell: \mathbb{R} \rightarrow \mathbb{R}_+$ can be written $\ell(v) = \ell_e(v) + \ell_o(v)$, where $\ell_e(v) =\frac{1}{2}(\ell(v) + \ell(-v))$ is the even part and the odd part is $\ell_o(v)= \frac{1}{2} (\ell(v) - \ell(-v))$. $\ell$ is a linear odd loss if 
    \begin{align*}
        \ell_o(v) = bv
    \end{align*}
    for some constant $b$, i.e., $\ell(v)$ is of the form
    \begin{align*}
        \ell(v) = \ell_e(v) + bv.
    \end{align*}
\end{definition}
Since $\ell_o(v) = \frac{1}{2}(\ell(v)-\ell(-v)) = bv$, 
when $\ell$ is differentiable, taking derivatives shows this condition to be equivalent to gradient-symmetry (with $b = \frac{c}{2}$). However, LOLs are not assumed to be differentiable (nor strictly convex), making them a more general class.

Given this relationship, a natural question emerges: can the bias-variance decomposition for strictly convex gradient-symmetric losses be extended to this class? 
This is an especially pertinent question since some losses based on the hinge loss are linear odd, but not gradient-symmetric nor strictly convex~\citep{VanRooyen2015, Plessis2015}.
We begin to answer this question with the following theorem, which decomposes the expected risk into the expected margin and a target-independent component.

\begin{restatable}[Decomposition for the LOLs]{theorem}{lolDecomposition}\label{the:lol_decomp}
   Let $\ell$ be a LOL with odd part $\ell_o(v)=bv$ and even part $\ell_e(v) = \frac{1}{2}(\ell(v) + \ell(-v))$, then the expected risk decomposes as
   \begin{align*}
       \EXYsb{\EDb{\ell(Yf)}} = \EXb{bY^\ast f^\ast} + \EXb{\EDb{\ell_e(f)}},
   \end{align*}
   where $Y^\ast = \EYgivenXb{Y}$ and $f^\ast = \EDb{f}$.
\end{restatable}

This reveals that for LOLs (including gradient-symmetric losses) 
the expected risk can be decomposed into the expected margin and a function dependent only on the distribution of $f$.
This decomposition, an extension of the decomposition of the risk presented in~\citep{Patrini2016}, is not a bias-variance decomposition in a conventional sense: since $Y^\ast f^\ast$ can take any value in $\mathbb{R}$, there is no guarantee that the contribution of the first term is non-negative.

We can relate this decomposition to the bias-variance decomposition by noting that, for gradient-symmetric losses, the two terms can be written
\begin{align}
    \underbrace{\EXYsb{\ell(Yf^\ast)}}_{\textnormal{bias + noise}} = \EXb{b Y^\ast f^\ast + \ell_e(f^\ast)} \label{eq:lol_bias}
\end{align}
and 
\begin{align}
    \underbrace{\EXb{\EDb{\BregmanGen{\ell}{f}{f^\ast}}}}_{\textnormal{variance}} &= \EXb{\EDb{\ell_e(f)} - \ell_e(f^\ast)} \label{eq:lol_variance}.
\end{align}
The first,~\eqref{eq:lol_bias}, tells us that the expected risk is influenced by the target only via the expected margin. This simple form for the contribution of the target-dependent part of the risk is not dependent on the exact form of $\ell$ other than through the value of the constant $b$.%, and changes to the positive class probability $p$ influence the risk in a linear manner. 

In \eqref{eq:lol_variance}, we see that for gradient symmetric losses, the variance is the the difference between the two sides of Jensen's inequality (the Jensen gap).
This relationship highlights the importance of convexity; it is guaranteed that \eqref{eq:lol_variance} is non-negative if and only $\ell_e$ (and by extension $\ell$) is convex. 
We could use~\eqref{eq:lol_bias} and~\eqref{eq:lol_variance} as a bias-variance decomposition of sorts for the LOLs. However, when $\ell$ is is not strictly convex, we lose the guarantee that the variance term is non-zero for non-constant $f$, and when $\ell$ is non-convex, the variance can be negative. 

Finally, we note that the same exhaustive process for constructing LOLs in~\citep{Patrini2016} can be easily modified to give new gradient-symmetric losses: For any differentiable even function $\ell_e$, we can construct a gradient-symmetric loss as $\ell_e(v) - \frac{c}{2}v$ for any constant $c$.

\section{CONCLUSION}
We have presented a bias-variance decomposition for a broad class of \emph{gradient-symmetric} margin losses.
%, as well as a corresponding ambiguity decomposition.
Unlike previous work, our decomposition neither requires interpretation of the model as a class probability estimator nor computation of the inverse link or minimum risk functions.
From this bias-variance decomposition, we have shown more general decompositions applicable to non-gradient-symmetric losses, giving decompositions applicable to the more general classes of strictly convex losses and LOLs. 
While we have not derived a true bias-variance decomposition for the LOLs, we have shown that it is possible to separate out the target-dependent component of the expected risk.
%We leave full characterization of the class of losses for which bias-variance decompositions exist to future work.  

 The framework that we have developed opens up many avenues for exploration of bias and variance in practical settings, including examination of the longstanding question of whether boosting algorithms are bias or variance reducing~\citep{Mease2008}, development of 
bias-variance decompositions for SVMs with the L2 loss~\citep{Lee2013}, and possible
application of bias-variance decompositions for alternative learning scenarios, such as positive/unlabelled classification~\citep{Plessis2015} and learning from label proportions~\citep{Quadrianto09, Patrini2016}.

\subsubsection*{Acknowledgements}
The authors gratefully acknowledge the support of the EPSRC LAMBDA project (EP/N035127/1).

\bibliography{bibliography}

\clearpage
\appendix

\thispagestyle{empty}

% For one-column format, uncomment the following:
\onecolumn \makesupplementtitle
% For two-column format, uncomment the following:
%\twocolumn[ \makesupplementtitle ]

\section{Proofs for Section~\ref{sec:bias_variance}}

\generalBV*
\begin{proof}
    Define the random variable $V=Yf(\vectorX; D)=Yf$, then by the definition of $B_\ell$,
    \begin{align*}
    \EDb{\BregmanGen{\ell}{Y f}{Y f^\ast}} &=
        \EDb{\BregmanGen{\ell}{V}{\EDb{V}}} \\
            &= \EDb{\ell(V) - \ell(\EDb{V}) - \ell'(\EDb{V})(V - \EDb{V})}\\
        &= \EDb{\ell(V)} - \ell(\EDb{V}) - \ell'(\EDb{V})( \EDb{V}- \EDb{V})\\
        &= \EDb{\ell(V)} - \ell(\EDb{V}).
    \end{align*}
    Rearranging gives
    \begin{align*}
        \EDb{\ell(V)} = \ell(\EDb{V}) + \EDb{\BregmanGen{\ell}{V}{\EDb{V}}}.
    \end{align*}
    Then, expanding $V$ from its definition,
    \begin{align*}
        &\EDb{\ell(Yf)} = \ell(\EDb{Yf}) + \EDb{\BregmanGen{\ell}{Yf}{\EDb{Yf}}}\\
        \Rightarrow~
        &\EDb{\ell(Yf)} = \ell(Yf^\ast) + \EDb{\BregmanGen{\ell}{Yf}{Yf^\ast}}.
    \end{align*}
    Finally, taking the expectation with respect to $\vectorX$ and $Y$ completes the proof.
\end{proof}

\gradientSymmetricProperty*
\begin{proof}
    ($\implies$)
     Assume that for all $u,v \in \mathbb{R}$, we have $\BregmanGen{\ell}{u}{v}= \BregmanGen{\ell}{-u}{-v}$, then by the expanding both sides using the definition of Bregman divergences,  
    \begin{align*}
    \ell(u) - \ell(v) - \ell'(v)(u-v) = \ell(-u) -\ell(-v) - \ell'(-v)(-u+v). 
    \end{align*}
    Taking the derivative of both sides with respect to $u$ gives
    \begin{align}
        &\ell'(u) - \ell'(v) = -\ell'(-u) +\ell'(-v) \nonumber\\
        \implies & \ell'(u) + \ell'(-u) = \ell'(-v) + \ell'(v). \label{eq:gradsym_constant}
    \end{align}
    Since~\eqref{eq:gradsym_constant} is true for all $u, v \in \mathbb{R}$, $\ell'(u) + \ell'(-u)$ must be constant. 
    
    ($\impliedby$) We assume $\ell$ is gradient-symmetric, i.e., $\ell'(v) + \ell'(-v) = c$ for some $c \in \mathbb{R}$. We can take the anti-derivatives of both sides to get $\ell(v) - \ell(-v) = cv$ (note that the constant of integration is necessarily zero, since at $v=0$, $\ell(v) - \ell(-v)=0$). Considering $\BregmanGen{\ell}{u}{v}$,
    \begin{align*}
        \BregmanGen{\ell}{u}{v} &=  \ell(u) - \ell(v) - \ell'(v)(u-v)\\
        &= \ell(u) - \ell(v) - \underbrace{(c - \ell'(-v))}_{=\ell'(v)}(u-v)\\
        &= \ell(u) - \ell(v) -cu + cv + \ell'(-v)(u-v)\\
        &= \underbrace{\ell(-u) +cu }_{=\ell(u)} - \underbrace{(\ell(-v) +cv )}_{=\ell(v)} -cu +cv + \ell'(-v)(u-v)\\
        &= \ell(-u) +cu  - \ell(-v) -cv  -cu +cv - \ell'(-v)(-u-(-v)) \\
        &= \ell(-u) - \ell(-v) - \ell'(-v)(-u-(-v))\\ %\tag{Canceling terms}\\
          &= \BregmanGen{\ell}{-u}{-v}.
    \end{align*}
\end{proof}

\gradientSymmetricBV*

\begin{proof}
     $(\impliedby)$ 
    If $\ell$ is gradient-symmetric, then by Proposition~\ref{prop:grad_sym_condition}, $\BregmanGen{\ell}{u}{v} = \BregmanGen{\ell}{-u}{-v}$ for all $u,v \in \mathbb{R}$, and the sign of $Y$ is irrelevant in $\BregmanGen{\ell}{Yf}{Yf^\ast}$, therefore the variance term in ~\eqref{eq:gen_bv} can be expressed as $\EDb{\BregmanGen{\ell}{f}{f^\ast}}$ and the decomposition in~\eqref{eq:grad_sym_bv} holds. 
        
    $(\implies)$ We will show that if the decomposition holds for all choices of distributions of $\vectorX$, $Y$ and $D$, then for all choices of $u, v \in \mathbb{R}$,
    $\ell'(u) + \ell'(-u) = \ell'(v) + \ell'(v)$ and therefore $\ell$ is gradient-symmetric.
    
    Let $Y$ be the constant random variable $-1$ and $\vectorX=\vectorx$ for some constant $\vectorx$.
    For any $u, w \in \mathbb{R}$, we can define a distribution such that $\mathbb{P}(f(X, D) = u)= \mathbb{P}(f(X, D) = w) = \frac{1}{2}$. For this distribution, we have $f^\ast = \frac{1}{2} (u + w)$.
    
    Assuming the decomposition in~\eqref{eq:grad_sym_bv} holds, then by it and~\eqref{eq:gen_bv}, it must be true that
    \begin{align*}
        \EDb{\BregmanGen{\ell}{f}{f^\ast}} = \EDb{\BregmanGen{\ell}{-f}{-f^\ast}}.
    \end{align*}
    Expanding the expectations, 
    \begin{align*}
        \frac{1}{2} \left(\BregmanGen{\ell}{u}{\frac{1}{2}(u + w)} + \BregmanGen{\ell}{w}{\frac{1}{2}(u + w)} \right) &= \frac{1}{2} \left(\BregmanGen{\ell}{-u}{-\frac{1}{2}(u + w)} + \BregmanGen{\ell}{-w}{-\frac{1}{2}(u + w)} \right) \\
         \implies \BregmanGen{\ell}{u}{\frac{1}{2}(u + w)} + \BregmanGen{\ell}{w}{\frac{1}{2}(u + w)}  &=  \BregmanGen{\ell}{-u}{-\frac{1}{2}(u + w)} + \BregmanGen{\ell}{-w}{-\frac{1}{2}(u + w)}  
    \end{align*}
    For a fixed $w$, the above is true for all $u$ (after making the appropriate changes to the distribution of $D$). For this to be the case, the derivatives of both sides with respect to $u$ must be the same, i.e,
    \begin{align*}
        \frac{\partial}{\partial u} \left[ \BregmanGen{\ell}{u}{\frac{1}{2}(u + w)} + \BregmanGen{\ell}{w}{\frac{1}{2}(u + w)} \right] &= \frac{\partial}{\partial u} \left[ \BregmanGen{\ell}{-u}{-\frac{1}{2}(u + w)} + \BregmanGen{\ell}{-w}{-\frac{1}{2}(u + w)} \right]
    \end{align*} 
    We now proceed to simplify both sides. 
    Starting with the left-hand side, we find
    \begin{align}
        \frac{\partial}{\partial u} \left[ \BregmanGen{\ell}{u}{\frac{1}{2}(u + w)} + \BregmanGen{\ell}{w}{\frac{1}{2}(u + w)} \right]\hspace{-3cm}& \\
        &= 
        \frac{\partial}{\partial u} \big[\ell(u) - \ell(\nicefrac{1}{2}(u + w)) - \ell'(\nicefrac{1}{2}(u + w)) \Big(u - \nicefrac{1}{2}(u + w)\Big) \nonumber \\
         & \hspace{1.5cm} +\ell(w) - \ell(\nicefrac{1}{2}(u + w)) - \ell'(\nicefrac{1}{2}(u + w))\Big(w - \nicefrac{1}{2}(u + w)\Big)   \Big]\nonumber \\
         &= \frac{\partial}{\partial u} \big[\ell(u) - \ell(\nicefrac{1}{2}(u + w)) + \ell(w) -  \ell(\nicefrac{1}{2}(u + w)) \nonumber \\
         & \hspace{1cm}\underbrace{- \ell'(\nicefrac{1}{2}(u + w)) \Big(u - \nicefrac{1}{2}(u + w)\Big)  - \ell'(\nicefrac{1}{2}(u + w))\Big(w - \nicefrac{1}{2}(u + w)\Big)}_{=0}  \Big]\nonumber \\
         &= \frac{\partial}{\partial u} \left[ \ell(u) - 2 \ell(\nicefrac{1}{2}(u + w) ) \right]\nonumber \\
         &= \ell'(u) - \ell'(\nicefrac{1}{2}(u+w)).\label{eq:bregman_deriv_1}
    \end{align}
    Similarly, for the right-hand side we find
    \begin{align*}
            \frac{\partial}{\partial u} \left[ \BregmanGen{\ell}{-u}{-\frac{1}{2}(u + w)} + \BregmanGen{\ell}{-w}{-\frac{1}{2}(u + w)} \right]  & = -\ell'(-u) + \ell'(-\nicefrac{1}{2}(u+w))
    \end{align*}
    Combining the two, we get that for all $u \in \mathbb{R}$,
    \begin{align*}
        \ell'(u) - \ell'(\nicefrac{1}{2}(u+w)) = -\ell'(-u) + \ell'(-\nicefrac{1}{2}(u+w)).
    \end{align*}
    Or equivalently
    \begin{align*}
        \ell'(u) +  \ell'(-u) = \ell'(\nicefrac{1}{2}(u+w)) + \ell'(-\nicefrac{1}{2}(u+w)).
    \end{align*} 
    This argument holds for all $w$, so for any $v \in \mathbb{R}$, setting $w = 2v - u$ gives $v = \nicefrac{1}{2}(w + u)$ and therefore for all $u$ and $v$,
    \begin{align*}
        \ell'(u) + \ell'(-u) = \ell'(v) + \ell'(-v).
    \end{align*}
    This means that $\ell'(f) + \ell'(-f)$ is constant and therefore $\ell$ is gradient-symmetric.    
\end{proof}

\subsection{Proof of Buja's Decomposition}

In order to prove Theorem~\ref{the:buja_bv}, we first require the following lemma, which is a result from~\citep{Pfau2013}. 
\begin{lemma}~\label{lem:pfau_prob}
For any strictly convex differentiable function $\phi$, 
    \begin{align*}
     \EDb{\Bregman{p}{q}} = \Bregman{p}{q^\ast} + \EDb{\Bregman{q^\ast}{q}},
    \end{align*}
    where $q^\ast = [\phi']^{-1}\left( \EDb{\phi'(q)} \right)$
\end{lemma}
With this result, we proceed to proof of the theorem. 
\bujaBV*
\begin{proof}
 By Theorem~\ref{the:zhang},
    \begin{align*}
        L(p,f) = \minrisk(p) + \BregmanGen{-\minrisk}{p}{q}.
    \end{align*}
    So, by application of Lemma~\ref{lem:pfau_prob}, we have
    \begin{align*}
        \EXYsb{\EDb{L(p,f)}} 
        &= \EXb{\minrisk(p)} + \EXb{\EDb{\BregmanGen{-\minrisk}{p}{q}}}\\
        &= \EXb{\minrisk(p)} +  \EXb{\BregmanGen{-\minrisk}{p}{q^\ast} + \EDb{\BregmanGen{-\minrisk}{q^\ast}{q}}}.
    \end{align*}
    By rearrangement, we have 
    \begin{align*}
        \EXb{L(p, f) - \minrisk(p)} = \EXb{\BregmanGen{-\minrisk}{p}{q^\ast}} + \EXb{\EDb{\BregmanGen{-\minrisk}{q^\ast}{q}}}.
    \end{align*} 
    
\end{proof}
\negativeLoss*
\begin{proof}
Note that for a function $\phi(p)$, we can write the convex conjugate as $[\phi]^\ast(v)=v [\phi']^{-1}(v) - \phi([\phi']^{-1}(v))$ (see, for instance~\citep{Nielsen2010}). We can therefore write the convex conjugate of the negative minimum risk as
\begin{align*}
    \left(-L \right)^\ast (v) &= v [-\minrisk']^{-1}(v) + \minrisk\left([-\minrisk']^{-1}(v)\right).
\end{align*}
Since by Proposition~\ref{prop:grad_sym_optimal_link}, $[-\minrisk]'(p)= -c \psi(p)$, we have $[-\minrisk']^{-1}(v) = \psi^{-1}\left(-\frac{v}{c}\right)$. Additionally, we remind the reader that, by definition of the minimum risk and optimal link functions, $\minrisk(p) = p \ell(\psi(p)) + (1-p) \ell(-\psi(p))$. Putting this together,
\begin{align}
    \left[-L \right]^\ast (v) &= v [-\minrisk']^{-1}\,(v) + \minrisk\left([-\minrisk']^{-1}(v)\right) \nonumber \\
    &= v \, \psi^{-1} \left(-\frac{v}{c} \right) + \minrisk\left( \psi^{-1} \left(-\frac{v}{c} \right) \right)\nonumber \\
    &= v \, \psi^{-1} \left(-\frac{v}{c}\right) + \psi^{-1} \left(-\frac{v}{c} \right) \ell\left(-\frac{v}{c}\right) + \left( 1- \psi^{-1} \left(-\frac{v}{c}\right) \right) \ell\left(\frac{v}{c}\right)\nonumber \\
    &= \psi^{-1}\left(-\frac{v}{c} \right) \left(v + \ell\left(-\frac{v}{c}\right) -  \ell\left(\frac{v}{c}\right) \right) + \ell\left(\frac{v}{c} \right).\label{eq:conjugate_working}
\end{align}
We claim that the first term in the final expression is identically zero. Note that a necessary and sufficient condition for $v + \ell(-v/c) - \ell(v/c)=0$ for all $v \in \mathbb{R}$ is that $cv + \ell(-v) - \ell(v)=0$ for all $v \in \mathbb{R}$. The latter rearranges to $\ell(v) = \ell(-v) + cv$. For differentiable $\ell$, this is equivalent to the condition for gradient-symmetric, since
\begin{align*}
    &\ell'(v) + \ell'(-v) = c \\
     \iff& \ell(v) - \ell(-v) = cv + d,
\end{align*}
where $d$ can be shown to be zero by considering that $\ell(0) - \ell(-0) = 0 = 0c  + d$. Therefore, by~\eqref{eq:conjugate_working}, 
\begin{align*}
    \left[-\minrisk \right]^\ast (v) = \ell \left(\frac{v}{c} \right).
\end{align*}

\end{proof}

\begin{lemma}\label{lem:c_generator}
    For any Bregman generator $\phi$, and any $c\neq0$, if we define $\phi_c$ such that for all $v$, $\phi_c(v)=\phi(\frac{v}{c})$, then for all $u, v \in \mathbb{R}$, 
    \begin{align*}
        \BregmanGen{\phi_c}{u }{v} = \BregmanGen{\phi}{\frac{u}{c}}{\frac{v}{c}}.
    \end{align*}
\end{lemma}
\begin{proof}
    \begin{align*}
        \BregmanGen{\phi_c}{u}{v} &= \phi_c(u) - \phi_c(v) - \phi_c'(v)(u-v)\\
        &=  \phi(u/c) - \phi(v/c) - \frac{\phi'(v/c)}{c} (u-v)\\
        &=  \phi(u/c) - \phi(v/c) - \phi'(v/c) (u/c-v/c)\\
        &= \BregmanGen{\phi}{\frac{u}{c}}{\frac{v}{c}}.
    \end{align*}
\end{proof}

\dualTheorem*
    \begin{proof}
 We define $\ell_c(v) = \ell(\frac{v}{c})$ and note that by Theorem~\ref{prop:negative_loss} $\ell_c(v) = \ell(\frac{v}{c}) = [-\minrisk]^\ast(v)$, so
        \begin{align*}
        \BregmanGen{\ell}{u}{v} 
            & = \BregmanGen{\ell}{\frac{cu}{c}}{\frac{cv}{c}}\\
            &= \BregmanGen{\ell_{c}}{cu}{cv} \\
            &= \BregmanGen{[-\minrisk]^\ast}{cu}{cv} \\
            &= \BregmanGen{-\minrisk}{[-L']^{-1}(cv)}{[-L']^{-1}(cu)}
        \end{align*}
        
    \end{proof}
 
\section{Proofs for Section~\ref{sec:ambiguity}}~\label{app:ambig}

\generalAmbiguity*
\begin{proof}
    The ambiguity decomposition can be derived as a special case of the bias-variance decomposition.  We consider an ensemble of models $f_1, \ldots, f_M$, 
    constant $X=\vectorx$ and constant $Y=y$. If we choose $D$ such that $\mathbb{P}(f = f_i) = \frac{1}{M}$, then Theorem~\ref{the:margin_variance_decomposition} gives
    \begin{align*}
        \EDb{\ell(Yf)} &= \ell(Yf^\ast) + \EDb{\BregmanGen{\ell}{f}{f^\ast}}\\
        \implies \averagei \ell(y f_i) &= \ell(y\fbar) + \averagei \BregmanGen{\ell}{f_i}{\fbar}.
    \end{align*}
    Rearranging completes the proof.
   
\end{proof}

\gradientSymmetricAmbiguity*
\begin{proof}
    ($\impliedby$) 
    Since the ambiguity decomposition can be formulated as a special case of the bias-variance decomposition and gradient-symmetry implies that the bias-variance decomposition, gradient-symmetry also implies the ambiguity decomposition.
%     Let $V_i = yf_i$ and $\Vbar= y \fbar$, then
%     \begin{align*}
%         \averagei \BregmanGen{\ell}{V_i}{\Vbar} 
%             &= \averagei \ell(V_i) - \ell(\Vbar) - \averagei \ell'(\Vbar)(V_i - \Vbar) \\
%             &= \averagei \ell(V_i) - \ell(\Vbar) -  \ell'(\Vbar)\left(\averagei V_i - \Vbar\right) \\
%             &= \averagei \ell(V_i) - \ell(\Vbar)
%     \end{align*}
%     Rearranging, we have
%     \begin{align}
%          \ell(\Vbar) = \averagei \ell(V_i) - \averagei(p) \BregmanGen{\ell}{V_i}{\Vbar}\nonumber\\
%          \implies 
%          \ell(y\fbar) = \averagei \ell(yf_i) - \averagei \BregmanGen{\ell}{yf_i}{y\fbar}%\label{eq:ambiguity_general}.
%     \end{align}
%     Since, by assumption, $\ell$ is gradient-symmetric, Proposition~\ref{prop:grad_sym_condition} gives
%     \begin{align*}
%          \ell(y\fbar) = \averagei \ell(yf_i) - \averagei \BregmanGen{\ell}{f_i}{\fbar}.
%     \end{align*}

    ($\implies$)
    The proof of this direction is identical to the proof of the same direction in Theorem~\ref{the:f_bv}, except we define $\fhat$ and $h$ as the members of an ensemble with $M=2$, rather than using them to define a probability distribution. We therefore omit the full details here.
\end{proof}

In Section~\ref{sec:proper}, we claim that the ambiguity decomposition for linearly combined models only exists for gradient-symmetric margin losses since \eqref{eq:gen_mean} holds if and only if the derivative of the negative minimum risk is an scaling of the optimal link function. Here, we explain in more detail why this is the case. The key observation here will be about when the ensemble average $\averagei f_i$ coincides with inverse link of the central probability prediction $\qbar = [-L]'^{-1} \left( [-L'](\psi^{-1}(f_i)) \right)$. To examine this, we first need the following result about quasi-arithmetic means.\footnote{Quasi-arithmetic means are combiner functions of the form $\kappa^{-1} \left( \averagei \kappa(v_i)\right)$, where $\kappa$ is a continuous monotonic function. }

\begin{theorem}[Adapted from \cite{Bullen2013handbook}, p.271, Sec. 4.1.2, Theorem 5]\label{the:gen_mean}
    Let $\mathcal{S}$ be a closed interval on the extended real line and let $\kappa: \mathcal{S} \rightarrow \mathbb{R} \cup \{-\infty, \infty\}$ and $\lambda: \mathcal{S} \rightarrow \mathbb{R} \cup \{-\infty, \infty\}$ be strictly monotonic continuous functions. Then
    \begin{align*}
        \kappa^{-1} \left( \averagei \kappa(v_i) \right) = \lambda^{-1} \left( \averagei \lambda(v_i) \right),
    \end{align*}
    for all $v_1, \ldots, v_M \in \mathcal{S}$,
    if and only if $\kappa(v) = a\lambda(v) + b$, for some $a, b \in \mathbb{R}$, $a \neq 0$
\end{theorem}

We note that in~\cite{Bullen2013handbook}, the theorem is for \emph{weighted} quasi-arithmetic means, however, the proof of \cite{Bullen2013handbook} still holds in both directions for the unweighted case. 

Applying this result to our ambiguity decomposition, we will find that this necessarily means that the optimal link is a scaled version of $-\minrisk'$, but first we require the following lemma.

\begin{lemma}\label{lem:b_neq_zero}
    If $\psi(p) = -a\minrisk'(p)+b$ for some constants $a,b \in \mathbb{R}$, then $b=0$.
    \begin{proof}
    Note that from~\citep{Masnadi2011} we have that, for all $p \in (0,1)$,
    \begin{align*}
        \psi(p) = -\psi(1-p)
    \end{align*}
    and since $\minrisk(p) = \minrisk(1-p)$ (\citep{Masnadi2011}), for all $p \in (0,1)$,
    \begin{align*}
        -\minrisk'(p) = \minrisk'(1-p)
    \end{align*}
       Using these two facts, we define the linear function such that $l(p) = ap + b$, so that $\psi = \ell \circ [-\minrisk]'$ and note that $-\minrisk'= l^{-1} \circ \psi$. The inverse of $\ell$ can be written $l^{-1}(v) = \frac{1}{a}v + \frac{b}{a}$. This gives
    \begin{align*}
       -\minrisk'(p) &= l^{-1} \circ \psi(p)\\
       &= \frac{1}{a} \psi(p) + \frac{b}{a} \\
       &= - \frac{1}{a} \psi(1-p) + \frac{b}{a},
    \end{align*}
    but also
    \begin{align*}
        -\minrisk'(p) &= \minrisk'(1-p)\\
            &= -l^{-1} \circ \psi(1-p)\\
            &= -\frac{1}{a} \psi(1-p) -\frac{b}{a}.
    \end{align*}
    The only way these two constraints are met is if $b=0$.
    \end{proof}
\end{lemma}

This lemma, combined with the previous theorem, gives the following result.

\combinerIsLinear*
    \begin{proof}
        \begin{align}
            \psi^{-1}(\fbar) = [-\minrisk']^{-1}\averagei[-\minrisk]'(\psi^{-1}(f_i)),
        \end{align}
        is equivalent to 
            \begin{align}
                \fbar = \psi \left( [-\minrisk']^{-1}\averagei[-\minrisk]'(\psi^{-1}(f_i))\right).
            \end{align}
        Because $\psi$ and $-\minrisk'$ are both continuous and invertible,\footnote{We know that $-\minrisk'$ is exists and is continuous because Theorem~\ref{the:buja_bv} requires a differentiable optimal link, which implies the minimum risk is differentiable~\citep{Zhang2004}, and because $-\minrisk$ is strictly convex, $-\minrisk'$ is continuous and strictly monotonic. } the function $\psi \circ [-\minrisk']^{-1}$ is continuous and invertible, and therefore it is a quasi-arithmetic mean and we can apply Theorem~\ref{the:gen_mean} to get 
        \begin{align*}
            \psi \left( [-\minrisk']^{-1}(f)\right) = af + b,
        \end{align*}
        for some constants $a, b$ with $a \neq 0$, and by Lemma~\ref{lem:b_neq_zero},  $b=0$. Defining $p = [-L']^{-1}(f)$, this gives
        \begin{align*}
            \psi(p) = a [-\minrisk]'(p) = -a \minrisk'(p).
        \end{align*}
    \end{proof}

\section{Proofs for Section~\ref{sec:grad_sym}}

Before presenting proof of Theorem~\ref{the:grad_sym_breg}, we prove some basic facts about convex functions and limits. 

\begin{lemma}\label{lem:deriv_bound}
Let $\phi:\mathbb{R} \rightarrow \mathbb{R}$ be a strictly convex differentiable function with $\lim_{v \rightarrow \infty} \phi(v) = c$ for some constant $c$. Then
\begin{align*}
    \lim_{v\rightarrow \infty} \phi'(v) =0
\end{align*}
\begin{proof}
    Since $\phi'$ is monotonically increasing, it either diverges to infinity or has the upper limit $\lim_{v \rightarrow \infty} \phi'(v) = \inf(\{c \in \mathbb{R}~:~c > \phi'(v),~~ \forall v \in \mathbb{R}\} )$. If it were the case that $\lim_{v\rightarrow\infty}\phi'(v)=\infty$ then $\lim_{v\rightarrow\infty}\phi(v)=\infty$, this contradicts our assumption, so $\lim_{v \rightarrow \infty} \phi'(v) $ must be finite. It can then be shown that if a limit exists, then it must be zero (see, for instance \cite{convex}).
\end{proof}
\end{lemma}

We also require the following lemma, the proof of which is adapted from~\cite{560909}.
\begin{lemma}\label{lem:fprimef_bound}
    Let $\phi:\mathbb{R} \rightarrow \mathbb{R}$ be a differentiable convex function with $\lim_{v \rightarrow \infty} \phi(v) = 0$, then $\lim_{v \rightarrow \infty} \phi'(v)v=0$

\end{lemma}

\begin{proof}
    
    We begin by noting two facts. Firstly, since $\phi$ is strictly convex and tending towards a finite limit, it must be strictly decreasing. Secondly, since $\lim_{v \rightarrow \infty}\phi(v)=0$, its derivative satisfies $\lim_{N \rightarrow \infty}\int_0^N \phi'(v) \mathrm{d}v = \lim_{N \rightarrow \infty} \phi(N) - \phi(0) = -\phi(0)$, so the derivative $\phi'$ has a finite integral over the positive reals.
    
    With these facts, we prove the contrapositive of the statement in our theorem. That is, we will show that if $v \phi'(v)$ does not converge to zero (which means there must exist $\epsilon>0$ such that for any $v_0$ there is always $v>v_0$ such that $|v \phi'(v)| > \epsilon$) 
    %(we derive this condition from the definition of limit and the fact that $\phi$ is necessarily monotonically decreasing due to its convexity)
    then $\int_0^\infty \phi'(v) \mathrm{d} v$ is not finite and therefore $\lim_{v\rightarrow \infty}\phi(v)\neq 0$. 
    
    Fixing some $\epsilon>0$, we define a sequence $\{v_n\}$ such that for each $n$ in the sequence, $v_n > 2 v_{n-1}$ and $|v_n \phi'(v_n)| > \epsilon$.  
    We have
    \begin{align*}
        \left|\int_{v_{n-1}}^{v_n} \phi'(v) \,\mathrm{d}v \right|\geq \left|\int_{v_{n-1}}^{v_n} \phi'(v_n) \, \mathrm{d}v \right| = \left|(v_n -v_{n-1})\phi'(v_n)\right|
    \end{align*}
    where we get the inequality due to the fact that $\phi'$ must be monotonically increasing towards $0$.
    Now, since $v_n - v_{n-1} > \frac{1}{2}v_n$, the assumption that $|v_n\phi'(v_n)| \geq \epsilon$ gives
    \begin{align*}
     \left|\int_{v_{n-1}}^{v_n} \phi'(v) \, \mathrm{d}v \right| > \left|\frac{1}{2}v_n\phi'(v_n)\right| \geq \frac{\epsilon}{2}.
    \end{align*}
    Since for all $N$, 
    \begin{align*}
        \left|\int_{0}^\infty \phi'(v) \, \mathrm{d}v\right| > \left|\sum_{i=1}^N \int_{v_{n-1}}^{v_n} \phi'(v) \, \mathrm{d}v\right| > N\frac{\epsilon}{2},
    \end{align*}
    the integral must be unbounded. 

\end{proof}

% \begin{lemma}\label{lem:limits}
%     For functions $p(v)$ and $q(v)$, $\lim_{v \rightarrow g} c\, p(v) + q(v)$ exists for all $c \in \mathbb{R}$ if and only if $\lim_{v \rightarrow g} p(v)$ and $\lim_{f \rightarrow g} q(v)$ both exist. 
%     \begin{proof}
%         If both $\lim_{v \rightarrow g} p(v)$ and $\lim_{v \rightarrow g} q(v)$ exist, then the existence of $\lim_{f \rightarrow g} c\, p(v) + q(v)$ is trivial. 
%         For the other direction, assume $\lim_{f \rightarrow g} c\, p(v) + q(v)$ exists for all $c$. Setting $c=0$ gives the existence of $\lim_{v \rightarrow g}q(v)$. Given the existence of $q(v)$, $\lim_{v_\rightarrow g} c p(v)$ must exist for all $c$ in order for the limit of the sum to exist, and given that $\lim_{v_\rightarrow g} c p(v)=c\,\lim_{v \rightarrow g} p(v)$, the limit of $p(v)$ must also exist. 
%     \end{proof}
% \end{lemma}

\gsAreBregman*
\begin{proof}
The general structure of this proof is as follows. First, we show if $\ell$ is expressible as a Bregman divergence, then the Bregman generator is necessarily gradient-symmetric. We then show that the Bregman generator must be $\ell$ (up to affine terms), and therefore $\ell$ is gradient-symmetric. Finally, we will prove the converse: if $\ell$ is gradient-symmetric, then it is necessarily expressible as a Bregman divergence with generator $\ell$.

\paragraph{Preliminaries: }   Note that
$g_y$ can take two values depending on the value of $y$. We denote the value it takes for positive $y$ as $\gpos$ and for negative $y$ as $\gneg$. We show that when \eqref{eq:conv_bregman} is satisfied, $\gneg = -\gpos$. 

When $y=+1$, we need our Bregman divergence to satisfy
\begin{align*}
    \ell(yf) = \ell(f) = \limpos \Bregman{f}{g},
\end{align*}
and similarly, when $y=-1$,
\begin{align*}
    \ell(yf) = \ell(-f) = \limneg \Bregman{f}{g}.
\end{align*}
Since $\ell$ is strictly convex with infimum zero, $\ell$ is zero at exactly one point (or limit) and that point is the minimum. Since Bregman divergences are zero if and only if their arguments are equal to each other, this means that when~\eqref{eq:conv_bregman} holds,
\begin{align}
    &\lim_{f \rightarrow \gpos} \Bregman{f}{\gpos} = 0\nonumber \\
    \Rightarrow &\lim_{f \rightarrow \gpos} \ell(f) = 0\nonumber\\
    \Rightarrow &\gpos = \argmin_{f \in \mathbb{R} \cup \{-\infty, \infty\}} \ell(f) \label{eq:gpos_def},
\end{align}
where $\ell(\infty) = \lim_{v \rightarrow \infty} \ell(v)$, and similarly for $\ell(-\infty)$.
We also have $\gneg = \argmin_{f \in \mathbb{R} \cup \{-\infty, \infty\}} \ell(-f)$ by the same reasoning. Combining these two facts, we have that $\gneg = -\gpos$. 

\paragraph{Proof that Bregman Divergence $\implies$ Gradient-Symmetry:}
\paragraph{$\phi$ is Gradient-Symmetric:} Using the symmetry of the labels that we have just established, we show that the generator $\phi$ must exhibit gradient symmetry (i.e., $\phi'(f) + \phi'(-f)$ must be constant). Observe that the margin loss has a symmetry, in that the loss is the same for the pair $f$ and $y$ as it is for $-f$ and $-y$, since trivially $\ell((-y)(-f)) = \ell(yf)$.
This gives that $\ell(f)$ must simultaneously satisfy both
\begin{align*}
    \ell(f) = \limpos \Bregman{f}{g}
\end{align*}
and 
\begin{align*}
    \ell(f) = \limneg \Bregman{-f}{g}.
\end{align*}
Expanding both of these using the definition of Bregman divergences and rearranging slightly, we find
\begin{align*}
    \ell(f) &= \lim_{g \rightarrow \gpos} \phi(f) - \phi'(g) f - \phi(g) + \phi'(g)g\\
    \ell(f) &= \lim_{g \rightarrow \gneg} \phi(-f) + \phi'(g) f - \phi(g) + \phi'(g)g
\end{align*}
Taking the derivatives of both expressions with respect to $f$, we have
\begin{align*}
    \ell'(f) &= \lim_{g \rightarrow \gpos} \phi'(f) - \phi'(g)\\
    \ell'(f) &= \lim_{g \rightarrow \gneg} -\phi'(-f) - \phi'(g).
\end{align*}
We can equate the two right-hand sides and rearrange to get
\begin{align*}
    \phi'(f) + \phi'(-f) =  \lim_{g \rightarrow \gneg} \phi'(g) - \lim_{g \rightarrow \gpos} \phi'(g).
\end{align*}
Since the term on the right-hand side is independent of $f$, the left-hand side must be constant with respect to $f$ and therefore $\phi$ is gradient-symmetric. 

To complete this direction of the proof, it now suffices to show that $\phi= \ell$ (up to affine terms).
%By convexity of $\phi$, we know that $\phi'$ is monotonically increasing and therefore the RHS is negative if $\gneg < \gpos$. Indeed, we can show that $\gpos>0$ and therefore $\gneg = -\gpos < \gpos$. To see this, note that if $f=\gpos$, by definition $\ell(yf)=0$. The fact that $\ell$ upper-bounds the zero-one, which is zero only for positive values of $f$, means that $f>0$. 

%The remainder of the proof is structured as follows: first, we establish that $\ell$ can be expressed as a Bregman divergence with generator $\phi(f) = \ell(-f)$ if and only if $\ell$ satisfies the conditions of being gradient-symmetric. To complete the proof, we establish that if $\ell$ can be expressed as a Bregman divergence, its generator \emph{must} be of the form $\phi(f) = \ell(f)$, up to affine terms---and therefore only $\ell$ which are gradient-symmetric are expressible as a Bregman divergence. 

\textbf{Necessity of $\phi(f) = \ell(f)$}: We assume that the loss can be expressed as a Bregman divergence and show that in this case, we must have $\phi(f) = \ell(f)$. 
If $\ell(yf)$ can be expressed as a Bregman divergence with generator $\phi$, then, taking $y=+1$, it must satisfy
\begin{align*}
    \ell(yf) = \ell(f) = \limpos \phi(f) - \phi'(g)f - \phi(g) + \phi'(g)g
\end{align*}
Note that, by definition, $\limpos \phi(g)=0$ and by Lemma~\ref{lem:deriv_bound}, $\limpos \phi'(g)=0$. Additionally, $\limpos \phi'(g)g = 0$ (trivially when $\gpos$ is finite and by Lemma~\ref{lem:fprimef_bound} when $\gpos$ is infinite). Therefore, taking the limit of these terms, we find
%In order for this expression to be defined, Lemma~\ref{lem:limits} tells us that $\limpos \, -\phi(g) + \phi'(g)g = c_1$ for some constant $c_1 \in \mathbb{R}$ and $\limpos \phi'(g)=0$ by definition of $\gpos$.
\begin{align}
    \ell(f) = \phi(f)\label{eq:margin_terms}
\end{align}
%Since Bregman generators are unique up to affine terms 
%$\phi(f)=\ell(f)$ is the only viable Bregman generator (up to affine terms).
\underline{This completes one direction of the proof}: If $\ell(yf)$ is expressible as a Bregman divergence, then $\ell$ is its Bregman generator, and $\ell$ is gradient-symmetric.

\paragraph{Proof that Gradient Symmetry $\implies$ Bregman divergence:} We now show that if $\ell$ is gradient-symmetric, it can be written as the limit of a Bregman divergence.
We do this by showing that 
\begin{align*}
    \limpos \BregmanGen{\ell}{f}{g} 
    &= \limpos [\ell(f) - \ell(g) - \ell'(g)(f-g)]\\
    &=\limpos [\ell(f) - \ell'(g)f - \ell(g) + \ell'(g)g]\\
     &= \ell(f) - \limpos [\underbrace{\ell'(g)f}_{\rightarrow 0} - \underbrace{\ell(g)}_{\rightarrow 0} - \underbrace{\ell'(g)g}_{\rightarrow 0}] \\
     &= \ell(f) = \ell(yf)|_{y=1}
\end{align*}
In the third line, we claim that three terms each converge to zero. The first of these is due to Lemma~\ref{lem:deriv_bound}, the second due to the condition we place on $\ell$ that its infimum is zero and the third due to Lemma~\ref{lem:fprimef_bound}. 

Now we do the same process for the negative limit $\limneg \BregmanGen{\ell}{f}{g}$. Since $\limpos \ell'(g)=0$ by Lemma~\ref{lem:deriv_bound}, the condition on the derivative gives $\limneg \ell'(g)=c$. Using this fact, along with the fact that gradient-symmetry implies $\ell(f) = \ell(-f) + cf$, we have

\begin{align*}
    \limneg \BregmanGen{\ell}{f}{g} 
    &= \limneg [\ell(f) - \ell'(g)f - \ell(g) + \ell'(g)g]\\
    &=  \underbrace{\ell(-f) + cf }_{= \ell(f)} - cf - \limneg \ell(g) + \limneg \ell'(g)g\\
    &=  \ell(-f) - \limneg \ell(g) + \limneg \ell'(g)g.
\end{align*}

Using the same identity on $\ell(g)$ and the identity $\ell'(g) = c- \ell'(-g)$,
\begin{align*}
    &= \ell(-f)  - \limneg \ell(-g) - \limneg cg  + \limneg (c - \ell'(-g)) g\\
    &= \ell(-f)  - \limpos \ell(g) - \limneg cg  + \limneg cg - \limpos \ell'(g)) g\\
    &= \ell(-f) - \limpos \ell'(g)g\\
    &= \ell(yf)|_{y=-1},
\end{align*}
where the last equality is due to Lemma~\ref{lem:fprimef_bound}. Combining the $y=+1$ case and the $y=-1$ case, we have that for gradient-symmetric losses, $\ell(yf) = \lim_{g \rightarrow g_y} \BregmanGen{\ell}{f}{g}$, completing the proof.

\end{proof}

\gradientSymmetricOptimal*
\begin{proof}
    If a loss is gradient-symmetric, it satisfies $\ell'(f)+ \ell'(-f)=c$ for some $c$, and if $c<0$ the loss $\widehat{\ell}(f)=\ell(f)/|c|$ satisfies $\widehat{\ell}'(f) + \widehat{\ell}'(-f)=-1$. This new loss has the same optimal link and is a canonical form loss. Since the minimum risk of this new loss is $1/|c|$ times the original minimum risk, 
    \begin{align*}
          \frac{1}{|c|} \minrisk(p) =\widehat{\minrisk}(p),
    \end{align*}
    where $\widehat{\minrisk}$ is the minimum risk of $\widehat{\ell}$. Since $\widehat{\ell}$ is in canonical form, by definition we have
    \begin{align*}
        -\widehat{\minrisk}'(p) = \psi(p) \implies -\frac{1}{|c|} \minrisk'(p) = \psi(p),
    \end{align*}
    which immediately gives $\minrisk'(p) = c\psi(p)$.
    
    %and the new loss shares its optimal link with the original loss $\ell$.
    %By \dw{Masnadi2011}, we know that for $\widehat{\ell}$, the optimal link and the gradient of the negative minimum are equal, so $\ell$ satisfies $c\minrisk'(p)= \psi(p)$.
    
    Reasoning in the other direction, if we start with a loss satisfying $\minrisk'(p)= c \psi(p)$, then it would share its optimal link with the loss $\widehat{\ell}$ and its minimum risk would be $|c|$ the minimum risk of $\widehat{\ell}$. Since $\widehat{\ell}$ is in canonical form (by virtue the negative derivative of its minimum risk being equal to its optimal link), it satisfies $\widehat{\ell}'(f) + \widehat{\ell}'(-f) = -1$. Therefore, $\ell(f) = |c| \widehat{\ell}(f)$ satisfies $\ell'(f) + \ell'(-f) = c$.
\end{proof}

\lolDecomposition*
\begin{proof}
If we look at the expected risk
\begin{align*}
    \EXYsb{\EDb{\ell(Yf)}} &= \EXYsb{\EDb{ \underbrace{\frac{1}{2}(\ell(Yf) + \ell(-Yf))}_{\textnormal{even part}} + \underbrace{\frac{1}{2}(\ell(Yf) - \ell(-Yf)}_{\textnormal{odd part}} }}\\
    &= \EXYsb{\EDb{\ell_e(Yf)  + \ell_o(Yf)}}\\
    &= \EXb{\EDb{\ell_e(f)}} + \EXYsb{\EDb{bYf}}\\
    &= \EXb{\EDb{\ell_e(f)}} + b Y^\ast f^\ast,
\end{align*}
\end{proof}

\section{Separating Bias and Noise}\label{sec:noise}

In the bias-variance decomposition in Theorem~\ref{the:f_bv}, the first term of the decomposition, $\EXYsb{\ell(Yf^\ast)}$, is the risk of the expected model, which is comprised of the noise, i.e., the systematic error that occurs due to uncertainty in the target; and the bias, i.e., the difference in loss between the predictor $f^\ast$ and the ideal predictor for the variable $Y$. It is not usually possible to experimentally measure the noise in a dataset, so for practical application of the decomposition, it makes sense to keep these quantities combined in a single term. However, from a theoretical standpoint, it is useful to further decompose this term. In this appendix, we discuss how this may be achieved.  

Theorem~\ref{the:zhang} shows that the excess risk can be written as a Bregman divergence, which is one of the key insights in constructing a bias-variance decomposition for these losses. 
However, rearranging the relationship in~\eqref{eq:excess_bregman}, we see that the pointwise risk for a fixed model $f$ can be written
\begin{align*}
    \EYgivenXb{\ell(Yf)} = L(p, f) =  \BregmanGen{-\minrisk}{p}{\psi^{-1}(f)} + \minrisk(p). 
\end{align*}
This decomposition of the pointwise risk has some striking similarities to the decompositions we have already seen: the expected error on the left-hand side is decomposed into two terms: the first term measures the separation between a point $\psi^{-1}(f)$ and the central point of a distribution $p=\EYsb{\Yzeroone}$, where $\Yzeroone=1$ when $Y=1$ and $\Yzeroone=0$ otherwise. More importantly, the observation allows us to break down the pointwise risk term and minimum risk terms in the following manner.

\begin{restatable}[Pointwise and Minimum Risks as Bregman Divergences]{theorem}{lpfBregman}~\label{the:lpfBregman}
    Let $\margin$ be a loss function with a differentiable minimum risk such that $\minrisk(0)> -\infty$ and invertible link function $\psi$. We have
    \begin{align}
        L(p, f) = \EYsb{\BregmanGen{-\minrisk}{\Yzeroone}{q}} + \minrisk(0),\label{eq:pointwiseBregman}
    \end{align}
    where $q = \optimallink^{-1}(f)$. Furthermore, for $p \in (0, 1)$,
    \begin{align}
        \minrisk(p) = \EYsb{\BregmanGen{-\minrisk}{\Yzeroone}{p}} + \minrisk(0).\label{eq:minriskBregman}
    \end{align}
    For fair losses (i.e., losses where $\minrisk(0)=\minrisk(1)=0$), the second term on the right-hand side vanishes. 
\end{restatable}
\begin{proof}
    We begin by showing the result for the minimum risk. To avoid confusion with signs, we use the notation $\phi = -\minrisk$ and consider the generator $\phi$.
    Using the definition of a Bregman divergence and the fact that $\Egenb{\Yzeroone}{\Yzeroone}=p$, we have
        \begin{align}
            \Egenb{Y}{\Bregman{\Yzeroone}{p}} &= \Egenb{Y}{ \phi(\Yzeroone) - \phi(p) - \phi'(p)(\Yzeroone-p)}\nonumber \\
            &= \Egenb{Y}{ \phi(\Yzeroone)} - \phi(p) - \phi'(p)(p-p) \nonumber \\
            &= \Egenb{Y}{ \phi(\Yzeroone)} - \phi(p) \nonumber \\
            &= p\phi(1) + (1-p)\phi(0) - \phi(p) \nonumber \\
            %&= p{\phi(1)} + (1-p) \phi(0) - \phi(p) \nonumber \\
            &= -p\minrisk(1) - (1-p) \minrisk(0) + \minrisk(p).\label{eq:minriskBregman2} 
        \end{align}
    For margin losses, $\minrisk(1)=\minrisk(0)$, so $p\minrisk(1) + (1-p) \minrisk(0)= \minrisk(0)$, and \eqref{eq:minriskBregman2} yields
    \begin{align*}
        \Egenb{Y}{\BregmanGen{-\minrisk}{\Yzeroone}{p}}= \minrisk(p) - \minrisk(0). 
    \end{align*}
    Rearranging gives the result
    \begin{align}
        \minrisk(p) = \Egenb{Y} {\BregmanGen{-\minrisk}{\Yzeroone}{p}}  + \minrisk(0) . 
    \end{align}
    Using this, we can prove the more general result for $L(p, f)$. Starting from the Bregman divergence for the excess risk ~\eqref{eq:excess_bregman} and applying ~\eqref{eq:minriskBregman}, we have 
    \begin{align*}
       L(p, f)    &= \BregmanGen{-\minrisk}{p}{q} + \minrisk(p)\\
       L(p, f) &= \BregmanGen{-\minrisk}{p}{q} + \Egenb{Y}{\BregmanGen{-\minrisk}{\Yzeroone}{p}} + \minrisk(0).
    \end{align*}
    Using the definition of Bregman divergences and again defining $\phi = -\minrisk$, and using $\EYsb{\BregmanGen{\phi}{\Yzeroone}{p}} = \EYsb{\phi(\Yzeroone)} - \phi(p)$, which we established as an intermediate step in~\eqref{eq:minriskBregman2}:
    \begin{align*}
        L(p,f) &= \phi(p) - \phi(q) - \phi'(q)(p - q) + \EYsb{\phi(\Yzeroone)} - \phi(p)  + \minrisk(0)\\
            &= -\phi(q) - \phi'(q)(p - q) + \EYsb{\phi(\Yzeroone)} + \minrisk(0) \\
            &= \EYsb{\phi(\Yzeroone) - \phi(q) - \phi'(q)(\Yzeroone -q)}  + \minrisk(0)\\
            &= \EYsb{\Bregman{\Yzeroone}{q}} + \minrisk(0).
    \end{align*}
\end{proof}

For a central model $f^\ast$, the bias and the noise can be separated using this result. 
Since the minimum risk can be written $\minrisk(p) = \EYsb{\BregmanGen{-\minrisk}{\Yzeroone}{p}} + \minrisk(0)$ and we already have an expression for the excess risk as a Bregman divergence, we can write
\begin{align*}
    \EXYsb{\ell(Yf^\ast)} &= \EXb{L(p, f^\ast)} \\
        &= \EXb{\minrisk(p) + \minrisk(p,f^\ast)  - \minrisk(p)}\\
        &= \minrisk(0) + \underbrace{\EXYsb{\BregmanGen{-\minrisk}{\Yzeroone}{p}} }_{\textnormal{noise}}+ \underbrace{\EXb{\BregmanGen{-\minrisk}{p}{\psi^{-1}(f^\ast)}}}_{\textnormal{bias}}.
\end{align*}

Note that $\minrisk(0)$ is a constant determined by the loss function, and \emph{fair losses}, including all those considered in this paper, is $\minrisk(0)=0$.
For gradient-symmetric losses, it is also possible to express this decomposition in terms of Bregman divergences with $\ell$ as the generator.
\begin{restatable}{theorem}{gradSymNoise}
    For a gradient-symmetric margin loss $\ell$ that satisfies $\minrisk(0)=0$, the noise can be written
   \begin{align*}
     \minrisk(p) = \EYsb{\lim_{g \rightarrow g_Y}\BregmanGen{\ell}{-\frac{\minrisk'(p)}{c}}{-\frac{g_Y}{{c}}}}
   \end{align*} 
\end{restatable}
\begin{proof}
    From Theorem~\ref{the:dual_connection}, we have for all $u, v \in \mathbb{R}$
    \begin{align*}
        \BregmanGen{\ell}{u}{v}  
            = \BregmanGen{-\minrisk}{[-L']^{-1}(cv)}{[-L']^{-1}(cu)},
    \end{align*} or equivalently, for all $u, v \in \mathbb{R}$
    \begin{align*}
        \BregmanGen{\ell}{\frac{u}{c}}{\frac{v}{c}}  
            = \BregmanGen{-\minrisk}{[-L']^{-1}(v)}{[-L']^{-1}(u)}
    \end{align*}
    Starting with the noise term, if we define the random variable $Y_\epsilon$ such that $\Yzeroone=1 \implies Y_\epsilon = 1-\epsilon$ and $\Yzeroone=0 \implies Y_\epsilon = \epsilon$,
    \begin{align*}
        \EYsb{\BregmanGen{-\minrisk}{\Yzeroone}{p}} 
        &= \EYsb{\lim_{\epsilon \rightarrow 0} \BregmanGen{-\minrisk}{Y_\epsilon}{p}}\\
        &= \EYsb{\lim_{\epsilon \rightarrow 0} \BregmanGen{\ell}{\frac{-\minrisk'(p)}{c}}{\frac{-\minrisk'(Y_\epsilon)}{c}}},
    \end{align*}
\end{proof}

\section{Relationship Between the Decompositions for Proper Scoring Rules and Gradient-Symmetric Losses}\label{app:coninciding_centroids}

In this appendix, we provide additional details about the relationship between the decompositions we present in Section~\ref{sec:bias_variance} and the decomposition in Theorem~\ref{the:buja_bv}. In Appendix~\ref{app:coinciding_centroids}, we show that the centroids in Theorem~\ref{the:margin_variance_decomposition} and Theorem~\ref{the:buja_bv} coincide if and only if the loss $\ell$ is gradient-symmetric. In Appendix~\ref{app:equiv}, we show that when both decompositions are defined, the bias and variance terms in Theorem~\ref{the:f_bv} and Theorem~\ref{the:buja_bv} are equivalent.

\subsection{Coinciding Centroids}\label{app:coinciding_centroids}
In this section, we prove that $\psi^{-1}(f^\ast)$ and $q^\ast$ coincide for a loss $\ell$ if and only if the loss is gradient-symmetric. 

We start with the following lemma, which is built around a result given by~\cite{3748031}.
\begin{lemma}\label{lem:cont_link}
    Let $\ell$ be a strictly convex differentiable margin loss, then $\psi$ is both continuous and invertible (with the range being $\mathbb{R}$ or a finite interval thereof).
\end{lemma}
\begin{proof}
Since $L(p, v) = p \ell(v) + (1-p) \ell(-v)$ and the sum of two strictly convex functions is also strictly convex, $L(p, v)$ is strictly convex in $v$. Furthermore, since margin losses (up to a constant factor) are upper bounds on the zero-one loss, both terms on the right hand side are positive.

We will show that $\psi$ is continuous around the arbitrary point $p_\infty \in (0,1)$.
Define the sequence $\{p_k\}$ with $p_k \in (0,1)$ for all $k$ and such that $p_k \rightarrow p_\infty$. Let $\{x_k\}$ be the sequence such that $x_k = \psi(p_k)$ and define $x_\infty = \psi(p_\infty)$. 

We will first show that all $x_k$ are contained in a compact set. We do this, as establishing this fact means that there is a convergent subsequence of $\{x_k\}$. We will show that this subsequence necessarily converges to $x_\infty$, and from the continuity of $L(p,v)$, this will give that $\psi$ is also continuous.

Since $\{p_k\}$ converges to $p_\infty \in (0, 1)$ we can choose $\epsilon>0$ such that all $p_k$, we have  $p_k \in (\epsilon, 1-\epsilon)$.
For all strictly convex functions on the real line either $\lim_{v \rightarrow \infty} \ell(v)= \infty$ or $\lim_{v \rightarrow \infty} \ell(-v) = \infty$. 
We will complete our proof assuming the latter case, though the former can be proved by simple modifications of the same argument. Assuming the latter, we now show for all $p \in (\epsilon, 1- \epsilon)$, that $\psi(p) \in \mathbb{R}$ is in a bounded region around $0$. Since $\ell(-v)$ tends to $\infty$, there exists $v_0$ such that for all $v> v_0$, $\ell(-v)> \frac{\ell(0)}{\epsilon}$ and therefore
\begin{align*}
    L(p, -v) 
        &= p\ell(-v) + (1-p) \ell(v)
        \\ &\geq p \ell(-v) 
        \\ &> p \frac{\ell(0)}{\epsilon} 
        \\ &> \ell(0) = L(p, 0).
\end{align*}
This means that for all $v > v_0$, we have $-v$ cannot be the minimizer of $L(p, \cdot)$ with $p \in (\epsilon, 1-\epsilon)$, so by definition of the optimal link function, $\psi(p) > - v_0$. In particular, for all $k$, $x_k = \psi(p_k) > -v_0$. 
Similarly for $v> v_0$, 
\begin{align*}
    L(p, v) &= p \ell(v) + (1-p) \ell(-v)\\
            & \geq (1-p) \ell(-v)\\
            & > \epsilon \ell(-v)\\
            & > \epsilon \frac{\ell(0)}{\epsilon}\\
            &= \ell(0) = L(p, 0).
\end{align*}
Hence, we can conclude that for $p \in (\epsilon, 1- \epsilon)$, $\psi(p) < v_0$.

Combining the two cases, we find that $x_k = \psi(p_k) \in (-v_0, v_0)$ for all $k$.
Since the sequence $\{x_k\}$ is all contained in a compact set, it necessarily has a convergent subsequence. We choose such a subsequence and call its limit $x_{\textnormal{acc}}$ (so called because it is an accumulation point of $\{x_k\}$).

Note that by definition of $\psi$ as the minimizer, for all $v \in \mathbb{R}$, we have 
\begin{align*}
L(p_k, x_k) = L(p_k, \psi(p_k)) \leq L(p_k, v).
\end{align*}

For our subsequence of $\{x_k\}$, by virtue of $L(\cdot, \cdot)$ being continuous, we have
\begin{align}
    \lim_{k \rightarrow \infty} L(p_k, x_k) =  L(p_\infty, x_{\textnormal{acc}}). \label{eq:compact2}
\end{align}
For the subsequence we also have
\begin{align}
   \lim_{k \rightarrow \infty} L(p_k, x_k) = \lim_{k \rightarrow \infty}L(p_k, \psi(p_k)) \leq \lim_{k \rightarrow \infty} L(p_k, v) = L(p_\infty, v),\label{eq:compact1}
\end{align}
%and for the entire sequence $\{p_k\}$, by continuity of $L$ with respect to $p$,
%\begin{align}
%    \lim_{k \rightarrow \infty} L(p_k, x_{\textnormal{lim}}) =  L(p_0, x_{\textnormal{lim}}). \label{eq:compact3}
%\end{align}
From~\eqref{eq:compact1} and~\eqref{eq:compact2}, we have for all $v \in \mathbb{R}$, that $L(p_\infty, x_{\textnormal{acc}}) \leq L(p_\infty, v)$, so $x_{\textnormal{acc}}$  is a minimizer of $L(p_0, \cdot)$. By strict convexity, this minimizer is unique so $x_{\textnormal{acc}}=x_\infty$.
%$\lim_{k \rightarrow \infty} L(p_0, x_{\textnormal{lim}}) = \lim_{k \rightarrow \infty} L(p_0, x_k)$, which, combined with~\ref{eq:compact1} gives that 
% $L(p_0, x_{\textnormal{lim}}) \leq L(p_0, v)$ for all $v$, and therefore $x_{\textnormal{lim}}$ is a minimizer of 

Consequently, we have that as $p \rightarrow p_\infty$, $\psi(p) \rightarrow \psi(p_\infty)$, so $\psi$ is continuous at $p_\infty$. Since $p_\infty$ is an arbitrary point on the interval $(0,1)$, $\psi$ is continuous on the entire interval. 

We now show that $\psi$ is strictly monotonic. For $v$ to be the minimizer of $L(p,v)$ for a specific $p$, it must be the case that 
\begin{align*}
    \frac{\partial}{\partial v} L(p,v)=0.
\end{align*}
Expanding and rearranging this, we find
\begin{align*}
    \frac{\partial}{\partial v} L(p,v)&=0\\
    \implies \frac{\partial}{\partial v} \left[ p \ell(v) + (1-p) \ell(-v)\right] &=0\\
    \implies p \ell'(v) - (1-p) \ell'(-v) &= 0\\
    \implies p &= \frac{\ell(-v)}{\ell'(v) + \ell'(-v)}.
\end{align*}
From this, we can see that $\psi$ must be strictly monotonic. If it were not, then by continuity, there would necessarily be $p_1\neq p_2$ such that $\psi(p_1)=\psi(p_2)$, but this would imply
\begin{align*}
    p_1 = \frac{\ell(-\psi(p_1))}{\ell'(\psi(p_1)) + \ell'(-\psi(p_1))} =\frac{\ell(-\psi(p_2))}{\ell'(\psi(p_2)) + \ell'(-\psi(p_2))}=p_2,
\end{align*}
contradicting the assumption that $p_1$ and $p_2$ are distinct. Therefore $\psi$ is continuous and strictly monotonic, and is therefore invertible (on some potentially restricted interval of the real line).  
%Since $\minrisk(p) = p \ell(\psi(p)) + (1-p)\ell(-\psi(p))$, if $\ell$ is strictly convex, then so is $-\minrisk$.  \dw{?}
\end{proof}

Note that for some losses, such as the squared margin loss $\ell(v) = (1- v)^2$, the optimal link function is not invertible on $\mathbb{R}$ and the result only holds if we restrict our model $f$ to have values only in some compact interval determined by $\ell$.

In Section~\ref{sec:bv_proper}, we claimed that the central model and $f^\ast$ and the model $\psi(q^\ast)$ coincide if and only the loss is gradient-symmetric. Here we formally state and prove the claim.

\begin{theorem}\label{the:coinciding_centroids}
    For a strictly convex differentiable margin loss $\ell$ where both Theorem~\ref{the:f_bv} and Theorem~\ref{the:buja_bv} hold, define the central model as
    \begin{align*}
        f^\ast = \EDb{f}
    \end{align*}
    and define the model corresponding to the central probability estimate as $\psi(q^\ast)$ where
    \begin{align*}
        q^\ast = [-\minrisk']^{-1} \left(\EDb{-\minrisk'(q)} \right)
    \end{align*}
    the two models coincide, i.e., $f^\ast = \psi(q^\ast)$ if and only if $\ell$ is gradient-symmetric.
\end{theorem}
\begin{proof}
    ($\implies$)
    Assume that $f^\ast = \psi(q^\ast)$, then by the definitions of $f^\ast$ and $\psi(q^\ast)$
    \begin{align*}
       \EDb{f} = \psi \circ [-\minrisk']^{-1} \left(\EDb{ -\minrisk' \circ \psi^{-1}(f)} \right)
    \end{align*}
    Since $\psi$ and $-\minrisk'$ are both monotonic and continuous (the former using Lemma~\ref{lem:cont_link}), so is their composition\footnote{We have not proved that $-\minrisk$ is differentiable, but note that if it is not $B_{-\minrisk}$ is not uniquely defined, and neither is the bias-variance decomposition of Theorem~\ref{the:buja_bv}}. By the application of Thereom~\ref{the:gen_mean}, this means that $\psi \circ [-\minrisk']^{-1}$ must be a linear function. Defining $l = \psi \circ [-\minrisk]'^{-1}$ where $\ell(v) = av + b$, we have $-\minrisk' = l^{-1} \circ \psi$ and by Lemma~\ref{lem:b_neq_zero}, $b=0$.  We therefore have that $\psi(p) = -a\minrisk'(p)$ for a constant $a$, and therefore by Proposition~\ref{prop:grad_sym_optimal_link}, $\ell$ is gradient-symmetric.

    ($\impliedby$)
    If $\ell$ is gradient-symmetric, then by Proposition~\ref{prop:grad_sym_optimal_link}, $-\minrisk'(p) = c \psi(p)$, and therefore taking the inverse functions of both sides $[-\minrisk']^{-1}(f) = \psi^{-1}(f/c) $. Applying these two facts to $\psi(q^\ast)$, we find
    \begin{align*}
    \psi(q^\ast) &= \psi \circ [-\minrisk]^{-1} \left( \EDb{[-\minrisk'] \circ \psi^{-1}(f)} \right) \\
    &=  \psi \circ [-\minrisk]^{-1} \left( \EDb{ cf} \right)\\
    &= \frac{\EDb{cf}}{c}
    = \EDb{f},
    \end{align*}
    so gradient-symmetry implies that $\psi(q^\ast)= f^\ast$
\end{proof}

\subsection{Equivalence of Terms Between Decompositions}\label{app:equiv}

\begin{proposition}[Equivalence of Bias Terms]\label{prop:equiv_bias}
For a strictly convex gradient-symmetric loss $\ell$ with invertible differentiable optimal link,
\begin{align*}
    \EYsb{\ell(Yf^\ast)} = \BregmanGen{-\minrisk}{p}{\psi^{-1}(f^\ast)} + \minrisk(p)
\end{align*}
\end{proposition}
\begin{proof}
 The result can be derived immediately from Theorem~\ref{the:zhang} and the definition of $L(p, f)$:
    \begin{align*}
        \EYsb{\ell(Yf^\ast)} &= L(p, f^\ast)\\
        &= L(p, f^\ast) - \minrisk(p) + \minrisk(p)\\
        &= \BregmanGen{-\minrisk}{p}{\psi^{-1}(f^\ast)} + \minrisk(p).
    \end{align*}
\end{proof}

\begin{proposition}[Equivalence of Variance Terms]
For a strictly convex gradient-symmetric loss $\ell$ with invertible differentiable optimal link, 
\begin{align*}
    \EDb{\BregmanGen{\ell}{f}{f^\ast}} &= \EDb{\BregmanGen{-\minrisk}{q^\ast}{q}}
\end{align*}
\end{proposition}
\begin{proof}
    We could use the Proposition~\ref{prop:equiv_bias} and the two bias-variance decompositions to show that the two variance terms are equal, but it is perhaps more instructive to show that the two terms are equivalent using facts that we know about gradient-symmetric losses.
    
    Let $h(v) = -cv$, then by Proposition~\ref{prop:grad_sym_optimal_link}, we have $-\minrisk' = h \circ \psi $. Developing this gives
    \begin{align*}
        &-\minrisk' = h \circ \psi\\
        \iff& h^{-1} \circ [-\minrisk'] = \psi\\
        \iff& \psi^{-1} = [-\minrisk']^{-1} \circ h\\
        \iff& [-\minrisk'] \circ \psi^{-1}= h.
    \end{align*}
    Remember that for gradient-symmetric losses, $\psi(q^\ast) = f^\ast $, so $q^\ast = \psi^{-1}(f^\ast)$, and we have
    \begin{align*}
        \EDb{\BregmanGen{-\minrisk}{q^\ast}{q}} 
        &= \EDb{\BregmanGen{-\minrisk^\ast}{[-\minrisk'(q)]}{-\minrisk'(q^\ast)}} \\
        &= \EDb{\BregmanGen{-\minrisk^\ast}{[-\minrisk'(\psi^{-1}(f))]}{-\minrisk'(\psi^{-1}(f^\ast)}} \\
        &= \EDb{\BregmanGen{-\minrisk^\ast}{-cf}{-cf^\ast}} \\
        &= \EDb{[-\minrisk^\ast](-cf) - [-\minrisk^\ast](-cf^\ast) - [-\minrisk^\ast]'(-cf^\ast)(f- f^\ast)}  \\
        &= \EDb{[-\minrisk^\ast](-cf) - [-\minrisk^\ast](-cf^\ast)},
    \end{align*}
    where the last line uses that $f^\ast = \EDb{f}$. Now, using Theorem~\ref{prop:negative_loss}, $[-\minrisk]^\ast(cv) = \ell(v) $, so
    \begin{align*}
        \EDb{\BregmanGen{-\minrisk}{q^\ast}{q}} &= \EDb{\ell(-f) - \ell(-f^\ast)}\\
        &=\EDb{\ell(-f) - \ell(-f^\ast) - \ell'(f^\ast)(f - f^\ast)}\\
        &= \EDb{\BregmanGen{\ell}{-f}{-f^\ast}}\\
        &= \EDb{\BregmanGen{\ell}{f}{f^\ast}},
    \end{align*}
    where the last line uses the fact that $\ell$ is gradient-symmetric, along with Proposition~\ref{prop:grad_sym_condition}.
\end{proof}

\section{Errata For AISTATS Camera-Ready Version}

This version of the supplementary material contains the following corrections versus the version in AISTATS Proceedings.

\begin{itemize}
    \item In proof of Proposition~\ref{prop:grad_sym_optimal_link}: ``its minimum risk would be $\frac{1}{|c|}$ the minimum risk" $\rightarrow$ ``its minimum risk would be $|c|$ the minimum risk" 
    \item In proof of Theorem~\ref{the:dual_connection}: $\BregmanGen{[-\minrisk]^\ast}{cv}{cu} \rightarrow            \BregmanGen{[-\minrisk]^\ast}{cu}{cv} $ 
    \item In proof of Proposition~\ref{prop:equiv_bias}: ``...follows immediately form Theorem~\ref{the:lpfBregman}" $\rightarrow$ ``...follows immediately from Theorem~\ref{the:zhang}"
\end{itemize}

\end{document}